%% file: main.tex
\documentclass{article}

\PassOptionsToPackage{numbers,sort&compress}{natbib}
\usepackage[preprint]{neurips_2026}

\usepackage[utf8]{inputenc}
\usepackage[T1]{fontenc}
\usepackage{float}
\usepackage{hyperref}
\usepackage{url}
\usepackage{booktabs}
\usepackage{amsfonts}
\usepackage{amsmath}
\usepackage{amssymb}
\usepackage{graphicx}
\usepackage{nicefrac}
\usepackage{microtype}
\usepackage{xcolor}
\usepackage{colortbl}
\usepackage{xspace}
\usepackage[most]{tcolorbox}
\usepackage{fontawesome5}
\usepackage{CJKutf8}
\usepackage{multirow}
\usepackage{subcaption}

\makeatletter
\renewcommand{\@notice}{}
\makeatother

\graphicspath{{figures/}}

\newcommand{\method}{\textsc{SeFi-Image}\xspace}
\definecolor{SeFiRowBlue}{RGB}{225,242,252}
\definecolor{ResourceBlue}{RGB}{0,105,200}
\newcommand{\sefitablefont}{\footnotesize\setlength{\tabcolsep}{4pt}\renewcommand{\arraystretch}{1.05}}

\newcommand{\resourcegithubicon}{%
  \raisebox{-0.22em}{\includegraphics[height=1.18em]{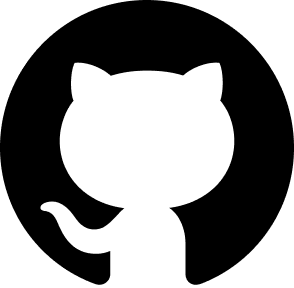}}%
}
\newcommand{\resourcepageicon}{%
  \raisebox{-0.12em}{\textcolor{ResourceBlue}{\faIcon{external-link-alt}}}%
}
\newcommand{\resourcemodelicon}{%
  \raisebox{-0.22em}{\includegraphics[height=1.18em]{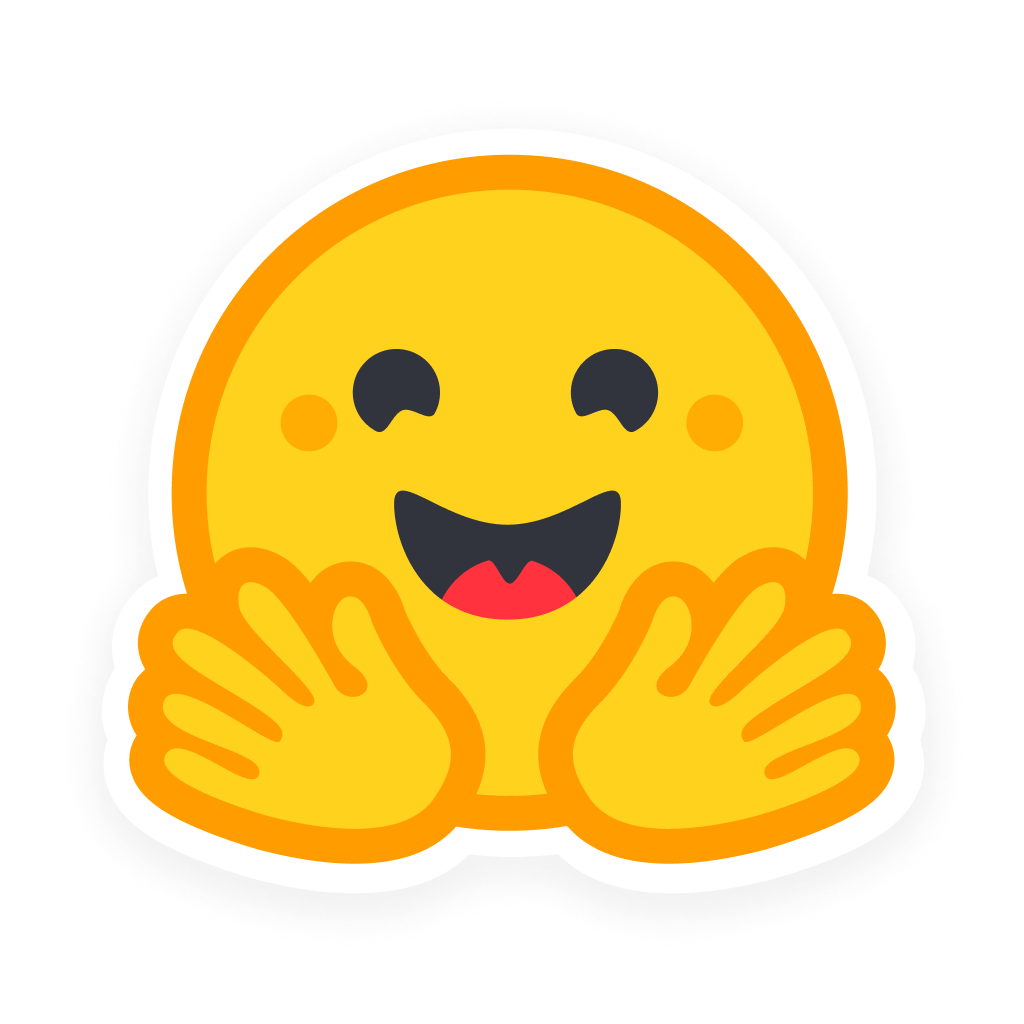}}%
}
\newcommand{\resourcelink}[1]{\textcolor{ResourceBlue}{\small\nolinkurl{#1}}}

\title{SeFi-Image: A Text-to-Image Foundation Model with Semantic-First Diffusion}

\author{SeFi Team}

\begin{document}

\maketitle

\input{chapters/00_abstract}

\begin{center}
\vspace{0.35em}
\begin{tcolorbox}[
  enhanced,
  width=0.94\textwidth,
  colback=white,
  colframe=white,
  boxrule=0pt,
  arc=3pt,
  left=20pt,
  right=20pt,
  top=8pt,
  bottom=8pt
]
\normalsize
\renewcommand{\arraystretch}{1.35}
\setlength{\tabcolsep}{0pt}
\begin{tabular}{@{}c@{\hspace{1.15em}}p{0.31\linewidth}@{\hspace{1.25em}}p{0.53\linewidth}@{}}
\resourcegithubicon & \textbf{GitHub} & \href{https://github.com/jmliu206/SeFi-Image}{\resourcelink{github.com/jmliu206/SeFi-Image}} \\
\resourcepageicon & \textbf{Project Page} & \href{https://jmliu206.github.io/sefi-web/}{\resourcelink{jmliu206.github.io/sefi-web/}} \\
\resourcemodelicon & \textbf{Hugging Face Model} & \href{https://huggingface.co/SeFi-Image}{\resourcelink{huggingface.co/SeFi-Image}} \\
\end{tabular}
\end{tcolorbox}
\end{center}

\begin{center}
\vspace{-0.15em}
\includegraphics[width=0.82\textwidth]{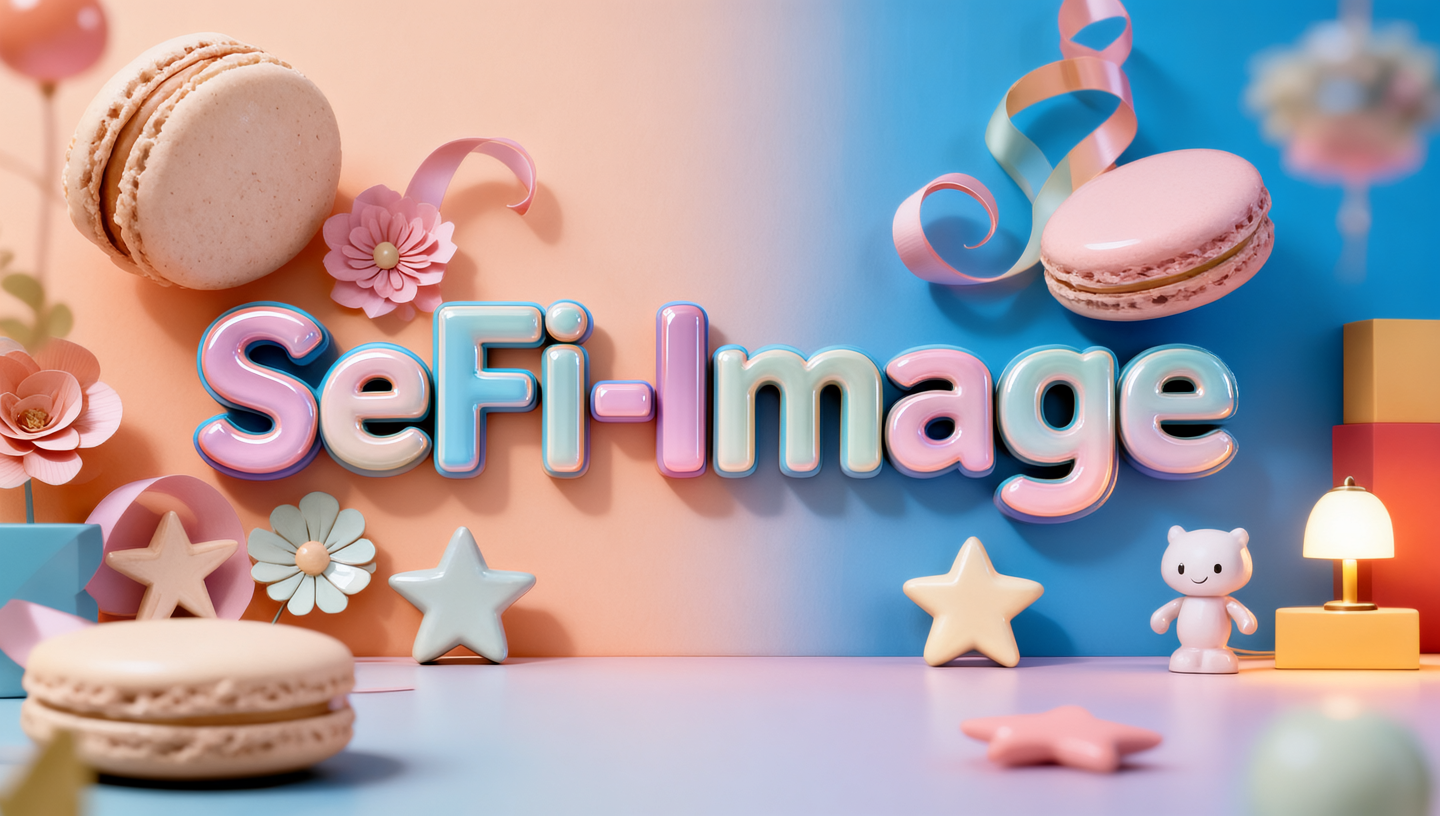}
\end{center}

\clearpage
\newgeometry{left=0.28in,right=0.28in,top=0.25in,bottom=0.50in}
\begin{figure}[p]
  \centering
  \makebox[\linewidth][c]{%
    \includegraphics[width=0.96\paperwidth,height=0.87\paperheight,keepaspectratio]{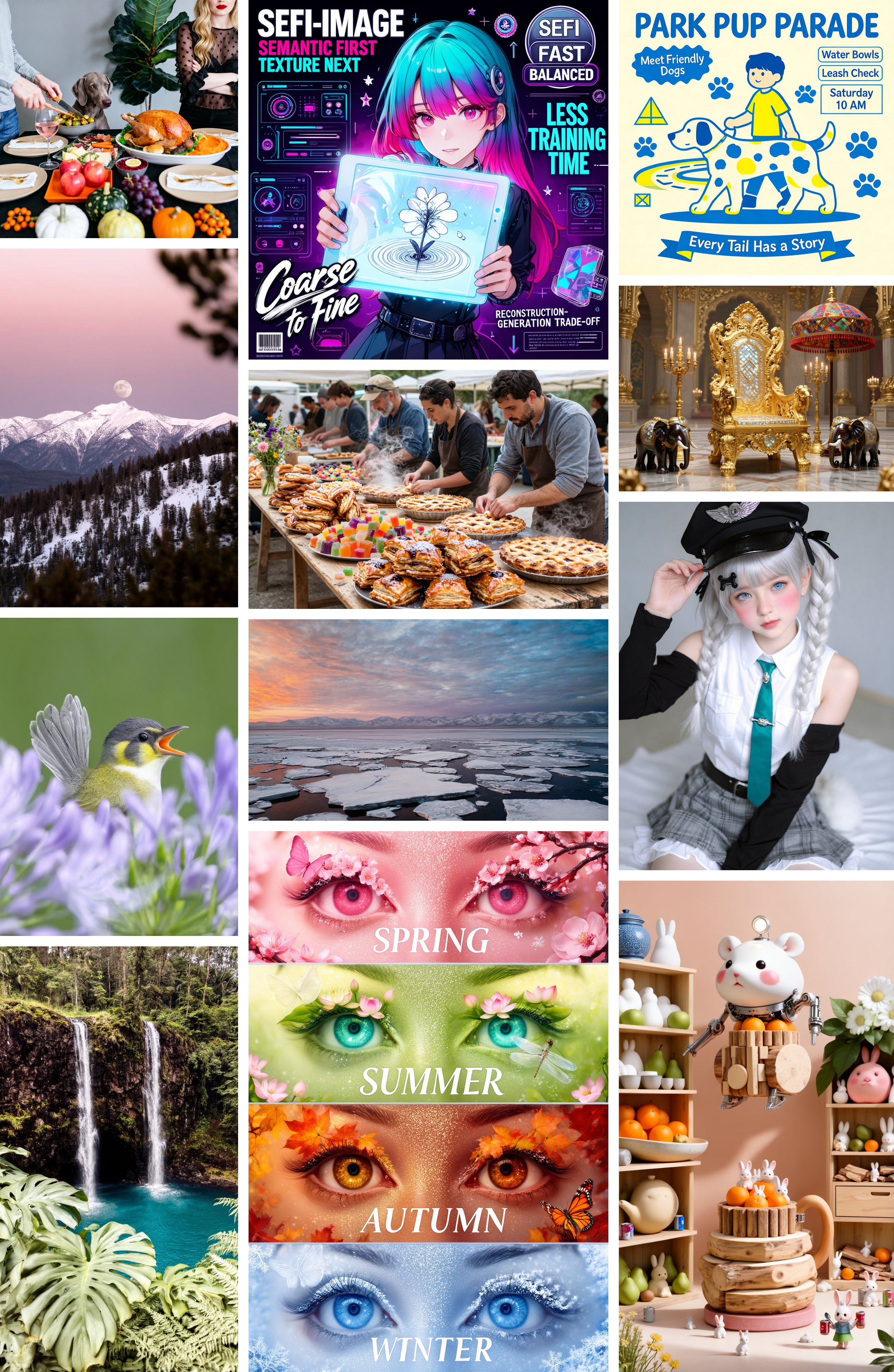}%
  }
  \caption{Images generated by \method.}
  \label{fig:teaser-g}
\end{figure}
\clearpage

\begin{figure}[p]
  \centering
  \makebox[\linewidth][c]{%
    \includegraphics[width=0.96\paperwidth,height=0.87\paperheight,keepaspectratio]{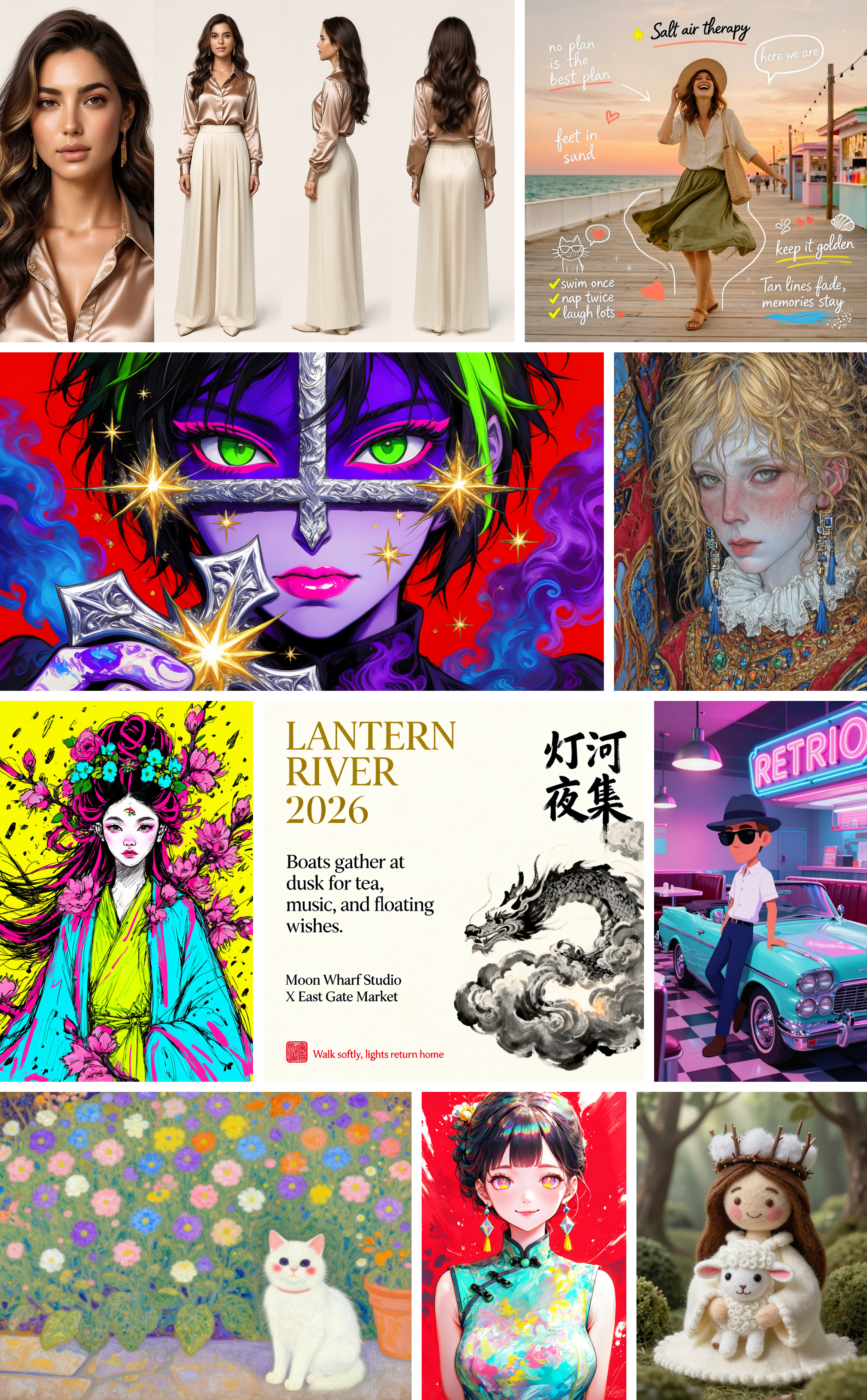}%
  }
  \caption{Additional images generated by \method.}
  \label{fig:teaser-c}
\end{figure}

\clearpage
\newgeometry{left=1.5in,right=1.5in,top=0.65in,bottom=0.65in}
\tableofcontents
\clearpage
\newgeometry{left=1.5in,right=1.5in,top=1in,bottom=1in}

\input{chapters/01_introduction}
\input{chapters/02_data}
\input{chapters/03_method}
\input{chapters/04_experiments}

\input{chapters/05_performance_evaluation}
\input{chapters/06_visualization}
\input{chapters/05_limitations}
\input{chapters/07_conclusion}

\section{Authors}
\textbf{Core Contributors:} Ruoyu Feng, Jinming Liu\\
\textbf{Contributors:} Yuqi Wang\footnote{Responsible for RL post-training.}, Xin Cheng, Boyuan Liu, Shanglin Li, Hanshen Zhu, Wenfeng Lin, Mingyu Guo, Xin Jin

\newpage
\input{chapters/references}

\appendix
\input{chapters/appendix}

\end{document}

%% file: chapters/00_abstract.tex

\begin{abstract}

Training image generation foundation models consumes substantial resources. Previous methods have attempted to leverage semantic guidance to accelerate the training process, yet their experiments were only conducted on simple datasets such as ImageNet, at low resolutions, and with small-scale models.
In this paper, we propose \method, a text-to-image foundation model built upon semantic-first diffusion, a novel latent diffusion modeling paradigm. We instantiate \method at three model scales, 1B, 2B, and 5B parameters, enabling systematic study of scaling behavior and flexible deployment under varying compute budgets. Notably, our largest 5B model was trained with merely 125K A800 GPU hours, corresponding to roughly 10-20\% of the training compute used by Z-Image. 
However, it achieves results comparable to or even superior to Qwen-Image and Z-Image. Despite this modest training compute, \method achieves strong performance on a wide range of benchmarks, including GenEval, DPG, LongTextBench, OneIG, and CVTG-2K. Moreover, we provide DMD2-distilled few-step turbo variants for each model scale to accommodate diverse hardware constraints and latency requirements.
We publicly release our code, weights and hope this work offers the community useful insights into semantic-guided diffusion modeling for T2I generation, while also providing practical and readily deployable model options.

\end{abstract}

%% file: chapters/01_introduction.tex

\section{Introduction}
\label{sec:introduction}

Foundational text-to-image generative models have advanced rapidly in recent years, achieving generating high quality images while faithfully following complex textual instructions \citep{rombach2022high,esser2024scaling,saharia2022photorealistic,gao2025seedream3,seedream2025seedream4,wu2025qwenimage,zimage2025,cao2025hunyuanimage,longcat2025}.
However, this progress has come with substantial training costs: even Z-Image, which explicitly emphasizes resource-friendly training, reports using 314K H800 GPU hours \citep{zimage2025}.

Recently, several methods have introduced semantic information from pretrained visual encoders to accelerate diffusion training. These include RAE~\citep{zheng2025diffusion} and VA-VAE~\citep{yu2025representation} for latent-space redesign or alignment \citep{zheng2025diffusion,yao2025reconstruction}, REPA for feature-level regularization \citep{yu2025representation}, REGLUE~\cite{petsangourakis2025reglue}, ReDi~\cite{kouzelis2026boosting} and REG~\cite{wu2026representation} for joint semantic-texture generation, and SFD for asynchronous semantic-texture modeling \citep{pan2025semantics}. These approaches have achieved significant convergence acceleration and improved FID on ImageNet 256$\times$256 class-conditional generation \citep{heusel2017gans,russakovsky2015imagenet}. However, these results have only been demonstrated with relatively small models (typically less than 1B) under class-conditional settings on toy datasets. Whether semantic guidance remains effective at larger model sizes and higher resolutions, and more importantly, whether it transfers well to the more practical text-to-image setting, remains an open question. Although several concurrent works~\citep{tong2026scaling,shi2025svg} have incorporated these methods into text-to-image models, none have attempted to build a truly state-of-the-art T2I foundation model on par with Qwen-Image and Z-Image. It also remains unclear whether such mechanisms offer benefits beyond faster convergence in the formal T2I training regime.

We propose \method, a text-to-image foundation model built upon Semantic-First Diffusion \citep{pan2025semantics}, a new latent diffusion modeling paradigm. Thanks to the semantic-first mechanism, \method achieves a superior reconstruction–generation trade-off: it operates in a VAE latent space with high reconstruction fidelity, yet still converges rapidly and attains strong final generation quality. We provide three model variants at 1B, 2B, and 5B parameters to accommodate diverse application requirements and hardware budgets. Even the smallest 1B model exhibits strong instruction-following capability. Notably, our largest 5B variant is trained with only 125K A800 GPU hours, corresponding to roughly 10–20\% of the training compute used by Z-Image. Despite this modest compute budget, \method achieves competitive and even stronger performance across a wide range of benchmarks, including GenEval, DPG, LongTextBench, OneIG, and CVTG-2K \citep{ghosh2023geneval,hu2024ella,geng2025xomni,chang2025oneig,tai2025complexvtg}. 
Moreover, we provide DMD2-distilled few-step turbo variants for each model scale to accommodate diverse hardware constraints and latency requirements. 
We hope this work offers the community useful insights into semantic-guided diffusion modeling for T2I generation, while also providing practical and readily deployable model options.


%% file: chapters/02_data.tex

\section{Data}
\label{sec:data}

\subsection{Pre-training}
\label{subsec:data-pretraining}


For pre-training, we use 450M internal image-text samples spanning a wide range of domains, dominated by natural images, together with 28M synthetic text-rendered image-text pairs.

\subsubsection{Image Caption}
A central component of our data pipeline is caption generation. For the 450M internal image pairs, we use Qwen3.5-2B \citep{qwen2026qwen35} to re-annotate all images, following three principles: \textit{accuracy, objectivity, and selective thoroughness}. Accuracy and objectivity mean that captions should faithfully describe what is visually present and avoid subjective or ambiguous language, so that each caption maps clearly to its image and provides clean supervision \citep{betker2023improving}.
Selective thoroughness requires the captioner to cover all important content with restraint, maximizing the learning signal from each sample and helping the model converge faster \citep{chen2026lens}, while avoiding hallucinations introduced by over-description.
This also bridges training and inference: since users typically provide detailed prompts for better results, a model trained on equally detailed captions faces less uncertainty at generation time. The captioning prompt is provided in Appendix~\ref{app:pretrain-caption-prompt}.

Meanwhile, our captions are bilingual, covering both Chinese and English, with each language provided in dense and short variants. During training, dense and short captions are sampled at a 4:1 ratio. This design exposes the model to dense captions more frequently, thereby providing richer supervisory signals, while still covering the short prompts that users may supply in practice. 

\subsubsection{Text-Rendered Synthetic Data}
\begin{figure}[t]
  \centering
  \begin{minipage}[t]{0.235\linewidth}
    \centering
    \includegraphics[width=\linewidth]{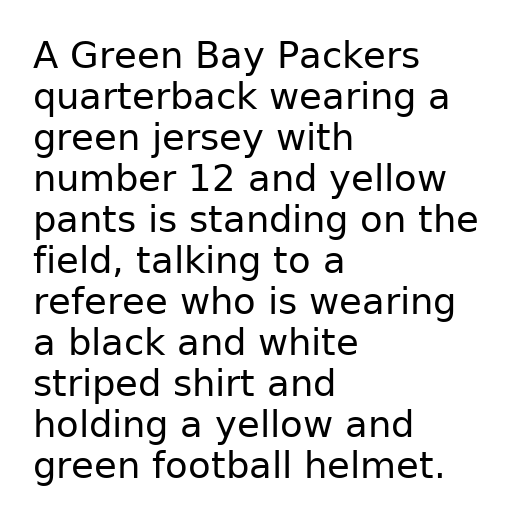}\\[-0.2em]
    {\small Part 1: dense English}
  \end{minipage}
  \hfill
  \begin{minipage}[t]{0.235\linewidth}
    \centering
    \includegraphics[width=\linewidth]{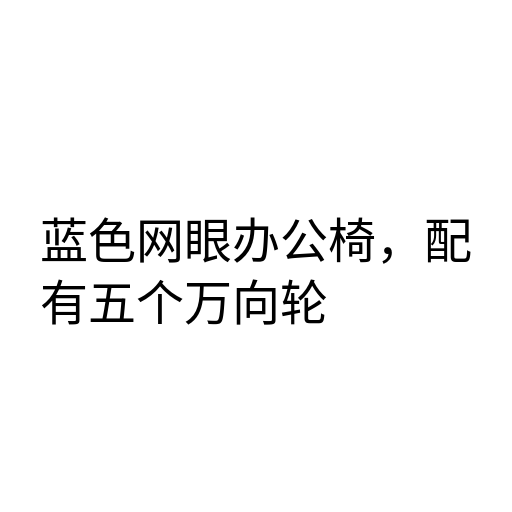}\\[-0.2em]
    {\small Part 1: short Chinese}
  \end{minipage}
  \hfill
  \begin{minipage}[t]{0.235\linewidth}
    \centering
    \includegraphics[width=\linewidth]{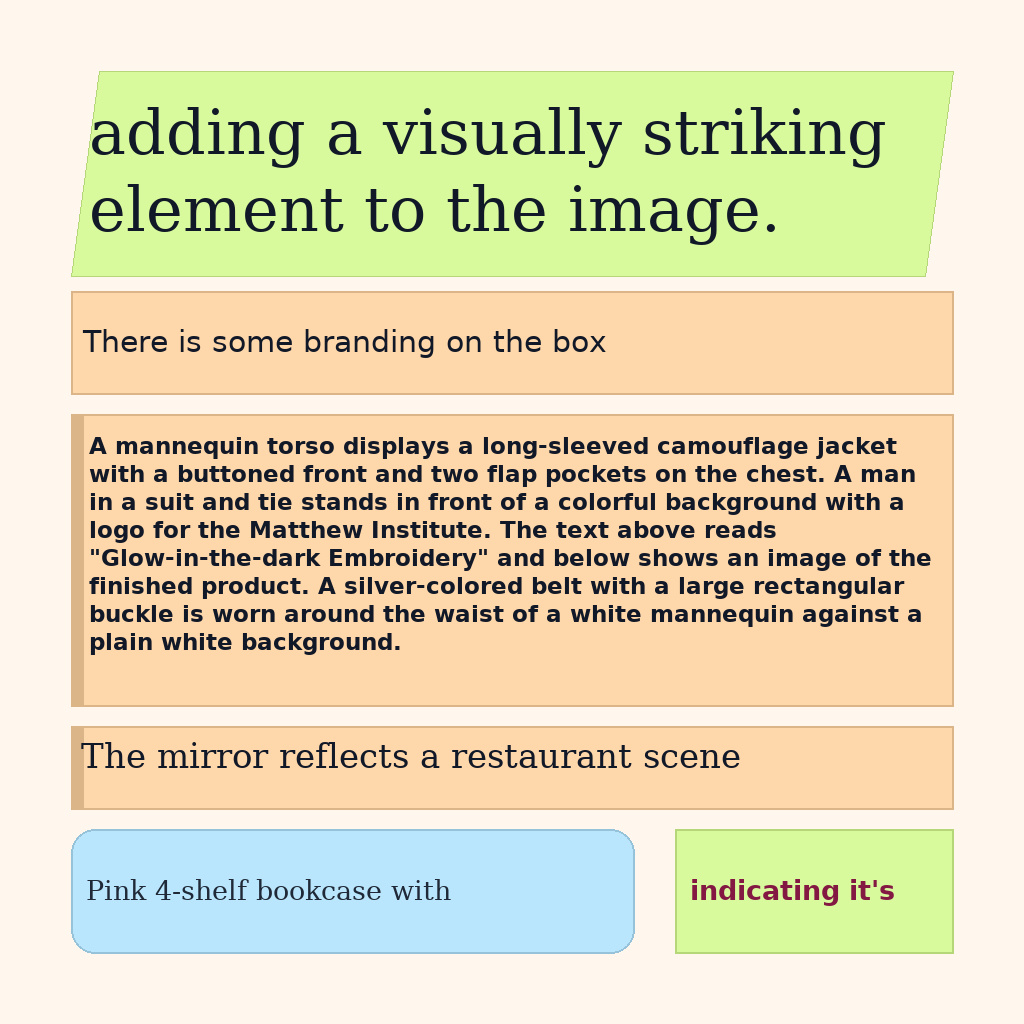}\\[-0.2em]
    {\small Part 2: structured layout}
  \end{minipage}
  \hfill
  \begin{minipage}[t]{0.235\linewidth}
    \centering
    \includegraphics[width=\linewidth]{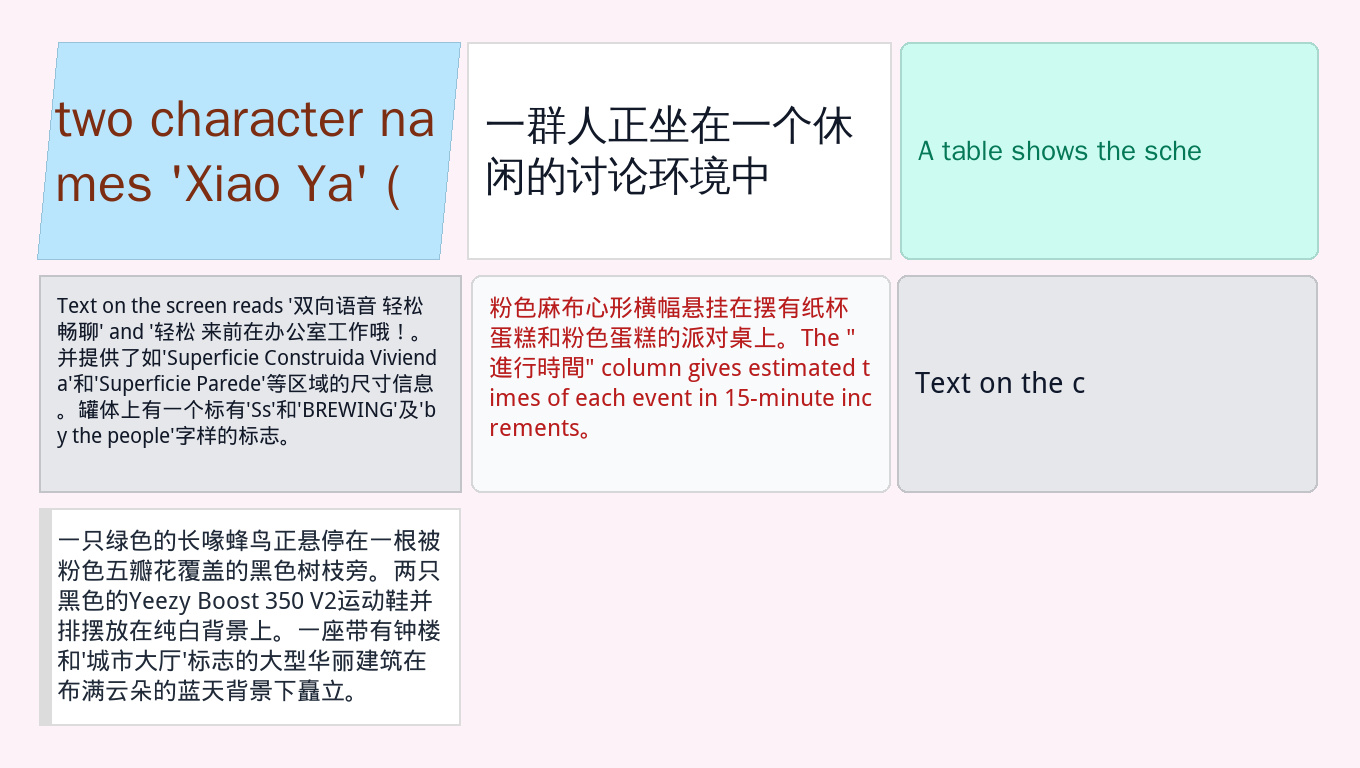}\\[-0.2em]
    {\small Part 2: mixed layout}
  \end{minipage}
  \caption{Examples from the two-part text-rendered synthetic data pipeline. Part 1 uses plain text blocks on simple backgrounds, while Part 2 introduces structured layouts with multiple text roles, colors, shapes, aspect ratios, and mixed-language content.}
  \label{fig:text-rendered-synthetic-examples}
\end{figure}

Accurately rendering text and arranging it according to specified layouts remains a key challenge for current image generation models. Historically, poor text rendering performance has been largely attributable to insufficient data and inaccurate captions. Collecting and filtering real-world text-rich images, and annotating them with precise captions, is non-trivial. In contrast, synthetic rendering can directly produce perfectly paired training data with exact ground-truth annotations. Recent works such as Z-Image~\cite{zimage2025} and Qwen-Image~\cite{wu2025qwenimage} have explored this direction by synthesizing text on solid-color backgrounds, compositing text onto complex scenes or paper-like textures, and performing ``cloze-style'' infilling on slides, aiming to cover diverse visual distributions and facilitate generalization to real-world imagery.
We argue that this can be approached more simply through a curriculum learning strategy. During pre-training, we focus on two core objectives: accurately rendering the text specified in the caption onto the image, and placing it at the designated position according to a given layout. Note that text rendering is essentially a strict one-to-one mapping; therefore, the semantic relevance between the rendered text and the image content is unimportant in the pre-training stage. What matters is ensuring sufficient diversity in the text itself. In the continual training and SFT stages, we introduce naturally distributed text-rich images, enabling the model to transfer its learned rendering capabilities to realistic text-in-image generation. Fig.~\ref{fig:text-rendered-synthetic-examples} shows examples of text-rendered synthetic data used during pre-training.

\textbf{Part~1: Plain text rendering.} This part generates images containing a single text block on a plain background. We use a PIL-based renderer to deterministically write text onto a 
$512\times512$ canvas. The text content is sampled from the same 450M recaptioned corpus used for pre-training, drawing from four caption variants (dense English, dense Chinese, short English, short Chinese) to form four balanced data buckets. Each rendered image is paired with a prompt of the form ``\texttt{The text in this image is ``\{text\}''.} ''(English) or ``\begin{CJK}{UTF8}{gbsn}\texttt{这张图片中的文字是``{text}''}\end{CJK}'' (Chinese), ensuring exact character-level alignment between caption and image. In total, this part produces 8M samples (2M per bucket).

\textbf{Part~2: Structured layout rendering.} This part extends text rendering to multi-block, multi-role layouts. We randomly generate diverse layout templates with varying aspect ratios ($1{:}1$, $4{:}3$, $16{:}9$, $3{:}4$, $9{:}16$, all at ${\sim}1024^2$ total pixels).
Each sample is composed of randomly generated template slots with varied layouts, colors, sizes, and shapes, filled with text rendered in diverse fonts, colors, and scales. The accompanying prompts describe the visible text content along with its position, color, and relative size. In total, we produce 20M samples (8M Chinese, 8M English, 4M mixed).

Across both parts, we enforce strict quality controls including character-level prompt-image alignment verification, overflow and bounding-box validation. 
The combined 28M text-rendered samples are mixed into the pre-training corpus to strengthen the model's text rendering accuracy, multi-block layout control, and reading-order awareness.

\begin{figure}[t]
  \centering
  \includegraphics[width=0.98\linewidth]{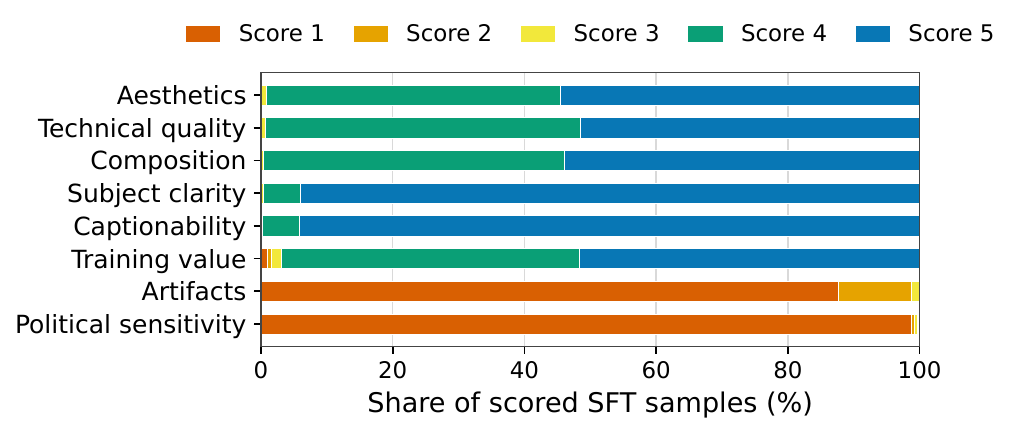}
  \caption{Distribution of VLM-based scores used by the SFT hard filtering gate. Scores are discrete values from 1 to 5. For artifacts and political sensitivity, lower scores indicate fewer issues.}
  \label{fig:sft-score-distribution}
\end{figure}

\subsection{Continual Training}
\label{subsec:data-continual-training}

For continual training, we curate a 9M image-text mixture comprising the Fine-T2I dataset \citep{ma2026finet2i} and internally collected data spanning diverse visual domains such as natural scenery, UI design, graphic design, anime, and so on. The data in this stage features higher quality and more challenging captions, which can substantially boost the model's instruction-following capability.

\subsection{Supervised Fine-tuning}
\label{subsec:data-sft}

After pre-training and continual training establish the foundational generation capability, we apply supervised fine-tuning (SFT) to further improve image aesthetics and instruction-following accuracy.

\paragraph{Data composition.}
The SFT stage uses around 650K high-quality images collected from several sources, including some open-source data, 200K Chinese text-rich images, and internally collected high-aesthetic samples. Compared with the pre-training corpus, this stage applies a much stricter quality bar. Each image is required to have strong aesthetic quality, clear composition, and a well-defined subject.

\paragraph{Annotation pipeline.}
Since SFT is more sensitive to caption quality than pre-training, we design a multi-stage annotation workflow based on proprietary VLMs. First, the VLM extracts structured metadata for each image, including semantic category (e.g., landscape, portrait, object, art, or poster), multilingual tags, safety attributes (NSFW, violence, and gore), watermark detection, OCR text with location and style information, and an initial quality assessment. The same pass also produces initial short and long captions in both Chinese and English. These captions are then refined to improve factual accuracy, level of detail, and language naturalness.

\paragraph{Quality scoring and filtering.}
Independently from caption generation, we score each image along multiple dimensions. The hard filtering gate uses core quality scores for aesthetics, technical quality, composition, subject clarity, captionability, and training value, together with artifact and political-sensitivity scores. In our rubric, technical quality measures low-level image fidelity such as sharpness, exposure, compression artifacts, resolution, and rendering stability; captionability measures whether the visible content can be objectively and faithfully described, rather than the quality of an existing caption; and training value summarizes the sample's overall usefulness for training after considering visual quality, semantic content, text rendering, artifacts, and dataset suitability. Auxiliary scores such as semantic richness, style strength, visual complexity, text-rendering quality, and commercial-design quality are retained for diagnostics, ranking, and distribution analysis rather than used as main hard thresholds. Fig.~\ref{fig:sft-score-distribution} shows the resulting score distributions for the dimensions used by the hard gate.

We then apply strict rule-based filtering. Images are rejected if they contain pornographic content, severe violence or gore, watermarks, heavy blur or corruption, unreadable key text, obvious artifacts, or politically sensitive content. The remaining images must also meet minimum thresholds on core scores, including aesthetics, technical quality, composition, subject clarity, captionability, and training value. We only keep images marked as core training samples, diversity supplements, text-layout samples, or style samples, while discarding borderline and low-value samples.

After deduplication, the final high-quality subset is exported together with refined captions and multilingual tags for training. During SFT, we use multiple text formats, including Chinese and English short captions, long captions, and tags, so that the model learns to respond to instructions with different levels of granularity.

%% file: chapters/03_method.tex

\section{Method}
\label{sec:method}

\subsection{Semantic-First Diffusion Modeling}
\label{subsec:modeling}

\begin{figure}[t]
  \centering
  \includegraphics[width=0.7\linewidth]{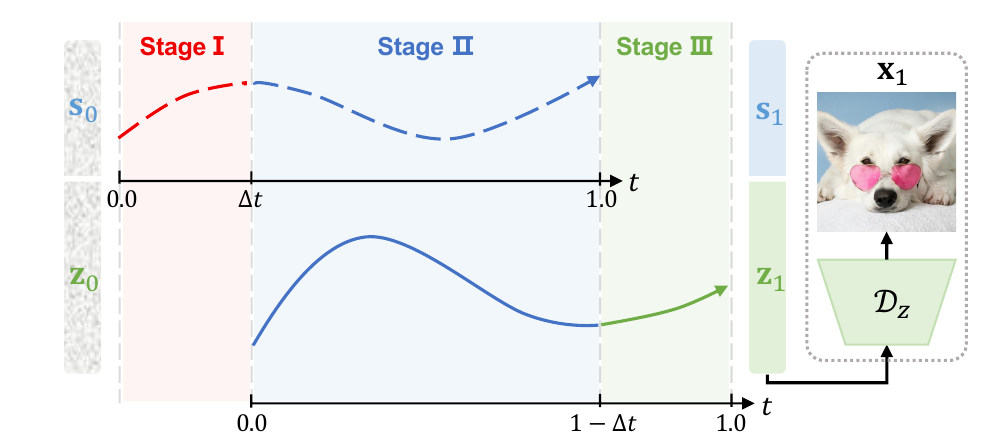}
  \caption{Illustration of Semantic-First Diffusion. The semantic latent is denoised ahead of the texture latent, providing a cleaner structural anchor for texture generation.}
  \label{fig:sfd-illustration}
\end{figure}


\method is built upon Semantic-First Diffusion (SFD) \citep{pan2025semantics}, a novel latent diffusion modeling paradigm. The key motivation of SFD is that image generation naturally follows a coarse-to-fine process: global semantics and object layout are usually established before high-frequency texture details. Conventional latent diffusion or flow-matching models denoise all information under a single shared timestep schedule, implicitly forcing semantic and texture factors to evolve synchronously. SFD instead separates these two factors along the diffusion timeline. Semantics are resolved slightly ahead of textures, so that texture generation is always conditioned on a cleaner semantic anchor, as illustrated in Figure~\ref{fig:sfd-illustration}.

\begin{figure}[h]
  \centering
  \includegraphics[width=\linewidth]{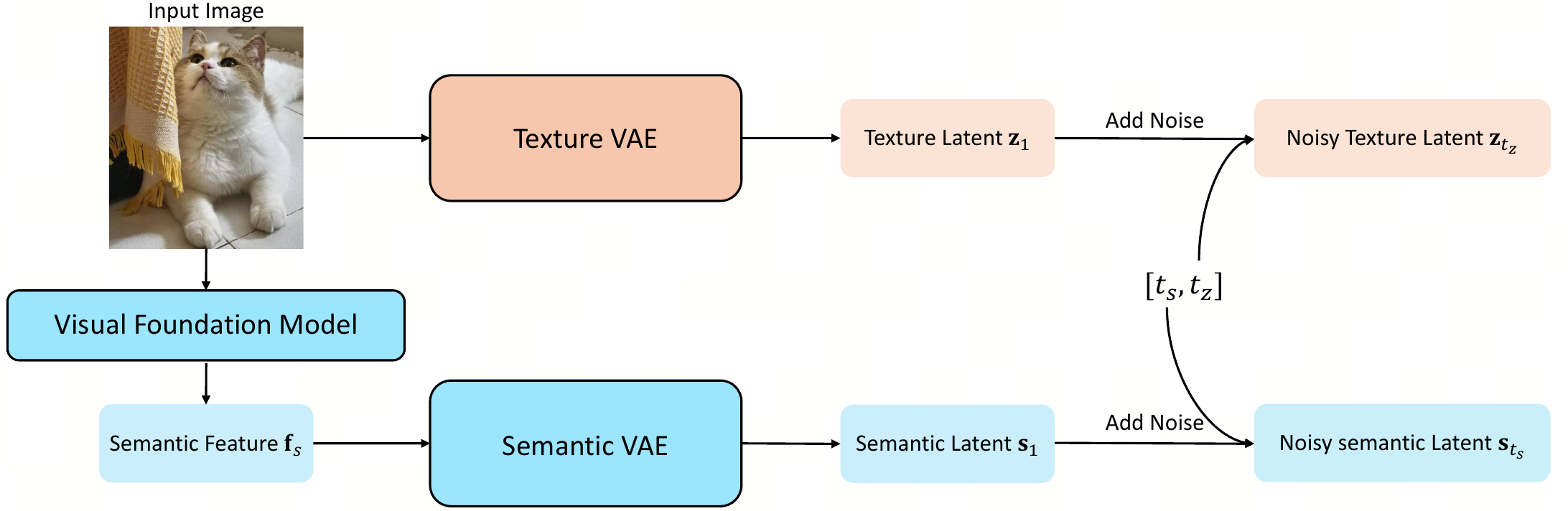}
  \caption{Construction of semantic and texture latents for SFD. The texture VAE encodes low-level reconstruction details into $\mathbf{z}_1$, while a visual foundation model and semantic VAE encode object identity, layout, and scene structure into $\mathbf{s}_1$. Independent noise is then added according to the texture and semantic timesteps.}
  \label{fig:sfd-latents-construction}
\end{figure}

\textbf{Composite latent construction.} As shown in Figure~\ref{fig:sfd-latents-construction}, for an image $\mathbf{x}$, we construct a composite latent from two sources. First, a frozen visual foundation model $\Phi$, instantiated as DINOv2-Large \citep{oquab2024dinov2}, extracts a semantic feature $\mathbf{f}_s=\Phi(\mathbf{x})$, which is then compressed by a semantic VAE encoder $\mathcal{E}_s$ into a compact semantic latent $\mathbf{s}_1=\mathcal{E}_s(\mathbf{f}_s)$. Second, a texture VAE encoder $\mathcal{E}_z$ maps the image directly to a texture latent $\mathbf{z}_1=\mathcal{E}_z(\mathbf{x})$. We then sample independent Gaussian noise $\mathbf{s}_0$ and $\mathbf{z}_0$, and follow flow matching~\citep{liuflow,esser2024scaling} paths to build noisy latent:
\begin{equation}
  \mathbf{s}_{t_s} = (1-t_s)\mathbf{s}_0 + t_s\mathbf{s}_1,\qquad
  \mathbf{z}_{t_z} = (1-t_z)\mathbf{z}_0 + t_z\mathbf{z}_1,
  \label{eq:sfd-flow-path}
\end{equation}
where $t_s,t_z\in[0,1]$ denote the semantic and texture timesteps, respectively. A timestep of $0$ corresponds to pure noise and $1$ to the clean latent.


\begin{figure}[t]
  \centering
  \includegraphics[width=\linewidth]{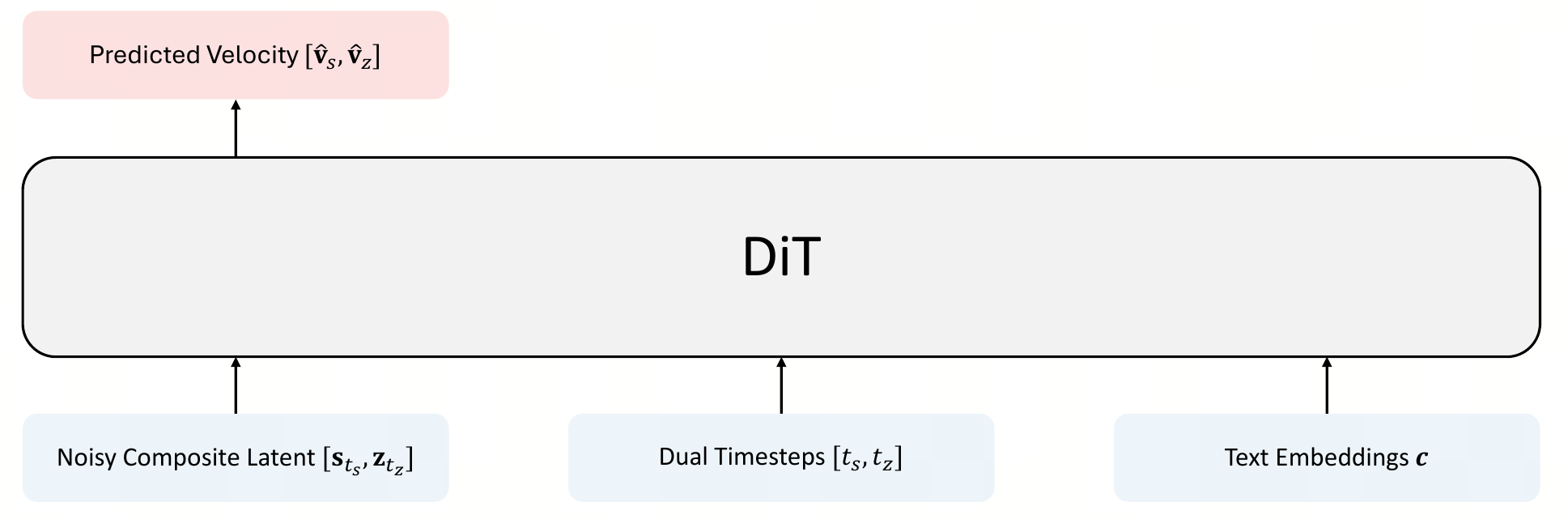}
  \caption{Overall framework of \method. The DiT takes the noisy composite latent, dual timestep embeddings, and text embeddings as input, and predicts velocity for both semantic and texture streams.}
  \label{fig:sfd-framework}
\end{figure}

\textbf{Distinct timesteps for semantics and textures.}
To model semantics and textures asynchronously with a fixed temporal offset $\Delta t$ while ensuring both timesteps remain within $[0,1]$, distinct timesteps $t_s$ and $t_z$ are assigned to the semantic and texture latents during training. Specifically, for each image, we first sample the semantic timestep $t_s$ from an extended interval, then derive the texture timestep $t_z$ by subtracting the offset $\Delta t$, and finally clamp both to $[0,1]$:
\begin{align}
  t_s &\sim \mathcal{U}(0,\, 1+\Delta t), \label{eq:sfd-ts-sample}\\
  t_z &= \max(0,\, t_s - \Delta t), \label{eq:sfd-tz-derive}\\
  t_s &= \min(t_s,\, 1), \label{eq:sfd-ts-clamp}
\end{align}
which ensures $t_s, t_z \in [0,1]$ and $t_s \geq t_z$. This guarantees the semantic latent experiences less noise corruption than the texture latent at each denoising step, thereby providing clearer structural guidance for texture denoising.


\textbf{Diffusion transformer with dual timesteps.} As shown in Figure~\ref{fig:sfd-framework}, the diffusion model adopts a Transformer backbone $\mathbf{v}_\theta(\cdot)$ that takes as input the noisy composite latent $[\mathbf{s}_{t_s}, \mathbf{z}_{t_z}]$ at different noise levels, two separate timesteps $[t_s, t_z]$, and the text condition $\mathbf{c}$:
\begin{equation}
  [\hat{\mathbf{v}}_s, \hat{\mathbf{v}}_z] = \mathbf{v}_\theta\left([\mathbf{s}_{t_s}, \mathbf{z}_{t_z}],\, [t_s, t_z],\, \mathbf{c}\right),
  \label{eq:sfd-vector-field}
\end{equation}
where $\hat{\mathbf{v}}_s$ and $\hat{\mathbf{v}}_z$ denote the predicted velocities of the semantic and texture components, respectively.

\textbf{Training objective.} The training objective combines velocity prediction losses for both semantic and texture latents:
\begin{equation}
  \mathcal{L}_{\mathrm{pred}}
  =
  \mathbb{E}_{\mathbf{s}_0,\mathbf{s}_1,\mathbf{z}_0,\mathbf{z}_1,t_s,t_z}
  \left[
    \left\|\hat{\mathbf{v}}_z - (\mathbf{z}_1-\mathbf{z}_0)\right\|^2
    +
    \beta
    \left\|\hat{\mathbf{v}}_s - (\mathbf{s}_1-\mathbf{s}_0)\right\|^2
  \right],
  \label{eq:sfd-loss}
\end{equation}
where $\mathbf{s}_0 \sim \mathcal{N}(0, I)$, $\mathbf{z}_0 \sim \mathcal{N}(0, I)$ are sampled from the prior, and $\beta$ is a weighting hyperparameter.

Additionally, the representation alignment loss from REPA~\citep{yu2025representation} is employed, which aligns the diffusion hidden states with pretrained vision encoder representations. Formally, it is defined as:
\begin{equation}
  \mathcal{L}_{\mathrm{REPA}}(\psi, \phi)
  :=
  -\mathbb{E}_{\mathbf{s}_{t_s},\mathbf{z}_{t_z},t_s,t_z}
  \left[
    \mathcal{L}_{\mathrm{sim}}\left(\mathbf{y}_*,\, h_\phi(\mathbf{h}_t)\right)
  \right],
  \label{eq:repa-loss}
\end{equation}
where $\mathbf{y}_* = f(\mathbf{x}_1)$ denotes the pretrained visual encoder output, $\mathbf{h}_t = f_\psi([\mathbf{s}_{t_s}, \mathbf{z}_{t_z}], [t_s, t_z])$ is the diffusion transformer encoder output, $h_\phi(\mathbf{h}_t)$ projects $\mathbf{h}_t$ through a trainable projection head, and $\mathcal{L}_{\mathrm{sim}}(\cdot,\cdot)$ is the alignment function. Notably, $\mathbf{y}_*$ corresponds to the semantic feature $\mathbf{f}_s$ input to the semantic VAE. 

The final objective is:
\begin{equation}
  \mathcal{L}_{\mathrm{total}} = \mathcal{L}_{\mathrm{pred}} + \lambda\, \mathcal{L}_{\mathrm{REPA}}.
  \label{eq:total-loss}
\end{equation}

\textbf{Three-phase denoising schedule.} During inference, SFD employs a three-phase asynchronous denoising schedule, as illustrated in Figure~\ref{fig:sfd-illustration}:
\begin{enumerate}
  \item \textbf{Semantic initialization}, where $t_s \in [0, \Delta t)$, $t_z = 0$: Only semantic latents are denoised to establish global structural guidance.
  \item \textbf{Asynchronous generation}, where $t_s \in [\Delta t, 1]$, $t_z \in [0, 1-\Delta t)$: Both semantic and texture latents are denoised jointly yet asynchronously, with semantics advancing slightly ahead to provide clearer structural guidance for texture generation.
  \item \textbf{Texture completion}, where $t_s = 1$, $t_z \in [1-\Delta t, 1]$: With semantic latents fully denoised, noisy texture latents continue to refine fine-grained details.
\end{enumerate}

Formally, two binary masks $\mathbf{M}_s \in \{0,1\}^{B \times C_s \times H \times W}$ and $\mathbf{M}_z \in \{0,1\}^{B \times C_z \times H \times W}$ are introduced to control the denoising updates of semantic and texture latents, respectively. According to the three-phase asynchronous denoising schedule, the masks $(\mathbf{M}_s, \mathbf{M}_z)$ are defined as:
\begin{equation}
  [\mathbf{M}_s, \mathbf{M}_z] =
  \begin{cases}
    [\mathbf{1}, \mathbf{0}], & t_s \in [0, \Delta t),\; t_z = 0,\\
    [\mathbf{1}, \mathbf{1}], & t_s \in [\Delta t, 1],\; t_z \in [0, 1-\Delta t),\\
    [\mathbf{0}, \mathbf{1}], & t_s = 1,\; t_z \in [1-\Delta t, 1],
  \end{cases}
  \label{eq:sfd-inference-masks}
\end{equation}
where $\mathbf{1}$ and $\mathbf{0}$ denote all-one and all-zero tensors with shapes matching $\mathbf{M}_s$ and $\mathbf{M}_z$, respectively. The masked velocity for updating is then computed as:
\begin{equation}
  \hat{\mathbf{v}} = \left[\mathbf{M}_s \odot \hat{\mathbf{v}}_s,\; \mathbf{M}_z \odot \hat{\mathbf{v}}_z\right],
  \label{eq:sfd-masked-velocity}
\end{equation}
where $\odot$ denotes element-wise multiplication. This mechanism explicitly controls which latents denoise at each phase, ensuring semantic latents denoise earlier to guide texture refinement continuously. By enabling asynchronous yet coordinated updates between semantic and texture latents, SFD achieves more stable optimization and naturally aligns with the coarse-to-fine generation paradigm of diffusion models.

Notably, while SFD extends the denoising timestep range by $\Delta t$, we proportionally increase the interval between successive steps, keeping the total number of diffusion steps fixed. Therefore, no additional denoising steps are required for inference. Upon completion, only the fully denoised texture latent $\mathbf{z}_1$ is decoded to the final image.

\subsection{Architecture}
\label{subsec:architecture}

\textbf{Texture VAE.} For the texture branch, we use a fine-tuned version of the FLUX.2 VAE \citep{flux2025frontier}. The original FLUX.2 VAE is already well-aligned with semantic structure and offers favorable learnability for generative model training. It uses 32 latent channels, twice the channel count of the FLUX.1 VAE \citep{flux2024}, providing a larger latent-space capacity that should, in principle, support stronger reconstruction quality. However, we observe that the posterior distribution of the FLUX.2 VAE exhibits relatively large variance, unlike earlier VAE designs that more directly prioritize reconstruction fidelity \citep{rombach2022high,esser2024scaling,flux2024}. We hypothesize that this stems from stronger KL regularization, which smooths the latent distribution and makes the texture latent space easier for diffusion models to learn.

Under the SFD modeling mechanism, however, texture latent generation is always
guided by a relatively cleaner semantic latent. This substantially reduces the
modeling burden of texture latent generation, making it possible to fine-tune the
Texture VAE more aggressively toward reconstruction quality without
substantially harming convergence or generative capacity. We therefore fine-tune
the FLUX.2 VAE to raise the reconstruction-generation trade-off of the overall
system. The fine-tuning objective is
\begin{equation}
  \mathcal{L}_{\mathrm{TexVAE}}
  =
  \mathcal{L}_{\mathrm{MSE}}
  +
  \lambda_{\mathrm{LPIPS}}\mathcal{L}_{\mathrm{LPIPS}}
  +
  \lambda_{\mathrm{KL}}\mathcal{L}_{\mathrm{KL}},
  \label{eq:texture-vae-loss}
\end{equation}
where $\mathcal{L}_{\mathrm{LPIPS}}$ denotes the perceptual similarity loss
\citep{zhang2018perceptual}. We set $\lambda_{\mathrm{LPIPS}}=0.1$ and
$\lambda_{\mathrm{KL}}=10^{-12}$. Since the resulting autoencoder is already
close to lossless compression, we do not introduce a GAN
loss. We train the texture VAE on the pre-training data. Since VAE reconstruction is primarily a low-level task and is less sensitive to the final training resolution, we apply 
$256\times256$ random crops as augmentation. The learning rate is set to $5\times10^{-5}$. The model is trained with a global batch size of 32 for 150K iterations on a single node of 8×A800 GPU, taking approximately 12 hours.

\begin{figure}[t]
  \centering
  \includegraphics[width=0.5\linewidth]{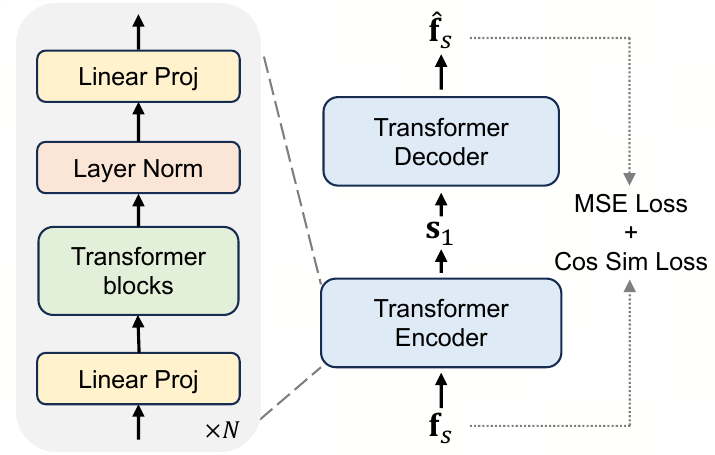}
  \caption{Illustration of the architecture of Semantic VAE (SemVAE).}
  \label{fig:architecture-semvae}
\end{figure}

\textbf{Semantic VAE.} Figure~\ref{fig:architecture-semvae} illustrates the Semantic VAE (SemVAE), which compresses high-dimensional visual foundation model features into a compact semantic latent space. Given an image $\mathbf{x}$, a frozen visual foundation model $\Phi$ extracts patch-level semantic features $\mathbf{f}_s = \Phi(\mathbf{x}) \in \mathbb{R}^{L \times C_{\mathrm{in}}}$, where $L$ is the number of flattened visual tokens and $C_{\mathrm{in}}$ is the feature dimension. The SemVAE encoder $\mathcal{E}_s$ first projects these features to the model hidden dimension, processes them through four Transformer blocks, and applies LayerNorm followed by a final linear projection:
\begin{equation}
  \mathbf{h}_s = \mathcal{E}_s(\mathbf{f}_s),\qquad
  \mathbf{h}_s \in \mathbb{R}^{L \times 2C_s},
  \label{eq:semvae-encoder}
\end{equation}
where $C_s$ is the semantic latent dimension. The channel dimension of $\mathbf{h}_s$ is split into the mean and variance parameters of a diagonal Gaussian posterior:
\begin{equation}
  \boldsymbol{\mu}_s, \boldsymbol{\sigma}_s^2
  =
  \mathbf{h}_s[:, :C_s],\; \mathbf{h}_s[:, C_s:],
  \label{eq:semvae-moments}
\end{equation}
and the semantic latent is obtained via the reparameterization trick:
\begin{equation}
  \mathbf{s}_1
  =
  \boldsymbol{\mu}_s + \boldsymbol{\sigma}_s \odot \boldsymbol{\epsilon},
  \qquad
  \boldsymbol{\epsilon} \sim \mathcal{N}(\mathbf{0}, \mathbf{I}).
  \label{eq:semvae-reparameterization}
\end{equation}
The SemVAE decoder $\mathcal{D}_s$ mirrors the encoder architecture and reconstructs the original VFM features from the sampled latent:
\begin{equation}
  \hat{\mathbf{f}}_s = \mathcal{D}_s(\mathbf{s}_1),\qquad
  \hat{\mathbf{f}}_s \in \mathbb{R}^{L \times C_{\mathrm{in}}}.
  \label{eq:semvae-decoder}
\end{equation}
This design preserves the spatial token layout of the VFM representation while compressing its channel capacity, allowing the diffusion model to operate on a compact semantic signal rather than directly modeling the full high-dimensional feature space. This also avoids the need to aggressively adjust the noise schedule~\citep{zheng2025diffusion}.

The SemVAE is trained independently before diffusion model training. During this stage, $\Phi$ is frozen and only $\mathcal{E}_s$ and $\mathcal{D}_s$ are optimized. Following the original SFD formulation \citep{pan2025semantics}, the training objective combines feature reconstruction, directional alignment, and latent regularization:
\begin{equation}
  \begin{aligned}
    \mathcal{L}_{\mathrm{MSE}}
    &=
    \left\|\hat{\mathbf{f}}_s - \mathbf{f}_s\right\|_2^2,\\
    \mathcal{L}_{\mathrm{cos}}
    &=
    1 -
    \frac{\hat{\mathbf{f}}_s \cdot \mathbf{f}_s}
    {\left\|\hat{\mathbf{f}}_s\right\|_2
     \left\|\mathbf{f}_s\right\|_2},\\
    \mathcal{L}_{\mathrm{KL}}
    &=
    D_{\mathrm{KL}}
    \left(
      q(\mathbf{s}_1 \mid \mathbf{f}_s)
      \,\middle\|\,
      \mathcal{N}(\mathbf{0}, \mathbf{I})
    \right).
  \end{aligned}
  \label{eq:semvae-loss-terms}
\end{equation}
The total SemVAE loss is:
\begin{equation}
  \mathcal{L}_{\mathrm{SemVAE}}
  =
  \mathcal{L}_{\mathrm{MSE}}
  +
  \mathcal{L}_{\mathrm{cos}}
  +
  \lambda_{\mathrm{KL}}\,\mathcal{L}_{\mathrm{KL}},
  \label{eq:semvae-loss}
\end{equation}
where $\lambda_{\mathrm{KL}} = 10^{-7}$. After training, the SemVAE encoder is frozen and used to produce semantic latents for SFD. The final image is decoded solely from the texture latent.

We train the SemVAE on the same data as the texture VAE, with a global batch size of 64, a learning rate of $5 \times 10^{-5}$, and a total of 1M iterations. Training takes approximately 48 hours on a single 8$\times$A800 GPU node.

\paragraph{Text Encoder.}
We use the LLM backbone of Qwen3-VL \citep{bai2025qwen3vl} as our text encoder, extracting hidden states from multiple layers and concatenating them as the text conditioning signal, similar to FLUX.2~\citep{flux2025frontier}. The choice of a large language model as the text encoder is motivated by its strong capabilities in understanding long and complex prompts, including multi-object relationships, counting, spatial reasoning, bilingual (Chinese and English) semantics, text rendering, rare concepts, and complex instruction following. For our 1B and 2B generation models, we adopt the LLM from Qwen3-VL-2B; for the 5B model, we scale up to Qwen3-VL-4B to provide richer text representations commensurate with the increased model capacity.

\paragraph{Transformer Architecture.}
We adopt a FLUX.2 [klein]~\citep{flux2025frontier} style DiT backbone that incorporating double-stream MMDiT blocks with single-stream blocks \citep{esser2024scaling,flux2024}. In the double-stream stage, visual tokens and text tokens are maintained as separate streams, each with its own normalization, modulation, and feed-forward layers, while cross-modal interaction is handled through joint attention. This separation is natural because image and text tokens are different modalities with different information properties. In the subsequent single-stream stage, the two token sequences are concatenated and processed by shared transformer layers, enabling deeper fusion and alignment. Detailed configurations of different model scales are illustrated in Table~\ref{tab:dit-architecture}.
Notably, two modifications adapt this architecture to SFD. First, because the visual token stream carries the channel-wise concatenation of semantic and texture latents, the input projection and output head are expanded accordingly. The transformer predicts a joint velocity field over the composite latent, which is split back into semantic and texture components for loss computation. Second, we replace the single timestep embedding with dual-timestep conditioning: the semantic and texture timesteps are embedded separately, concatenated, and used to modulate all transformer blocks. This makes the backbone aware of the asynchronous noise levels of the two streams at each denoising step.

\begin{table}[t]
  \centering
  \caption{DiT architecture configurations for the three \method model variants.}
  \label{tab:dit-architecture}
  \sefitablefont
  \setlength{\tabcolsep}{0.8pt}
  \begin{tabular}{@{}lcccccccccc@{}}
    \toprule
    \begin{tabular}[t]{@{}c@{}}Model\end{tabular} &
    \begin{tabular}[t]{@{}c@{}}Text\\encoder\end{tabular} &
    \begin{tabular}[t]{@{}c@{}}Inner\\dimension\end{tabular} &
    \begin{tabular}[t]{@{}c@{}}Attention\\heads\end{tabular} &
    \begin{tabular}[t]{@{}c@{}}Head\\dimension\end{tabular} &
    \begin{tabular}[t]{@{}c@{}}Double-stream\\blocks\end{tabular} &
    \begin{tabular}[t]{@{}c@{}}Single-stream\\blocks\end{tabular} &
    \begin{tabular}[t]{@{}c@{}}Total\\blocks\end{tabular} &
    \begin{tabular}[t]{@{}c@{}}Text\\dimension\end{tabular} &
    \begin{tabular}[t]{@{}c@{}}MLP\\dimension\end{tabular} &
    \begin{tabular}[t]{@{}c@{}}DiT\\parameters\end{tabular} \\
    \midrule
    1B & Qwen3-VL-2B & 2048 & 16 & 128 & 4 & 12 & 16 & 6144 & 6144 & $\sim$1.18B \\
    2B & Qwen3-VL-2B & 2560 & 20 & 128 & 4 & 16 & 20 & 6144 & 7680 & $\sim$2.18B \\
    5B & Qwen3-VL-4B & 3328 & 26 & 128 & 6 & 21 & 27 & 7680 & 9984 & $\sim$4.97B \\
    \bottomrule
  \end{tabular}
\end{table}

\section{Superiority of Semantic-First Diffusion}
\label{sec:experiments}


Before full-scale training, we conducted ablation experiments with a constrained dataset to verify the superiority of Semantic-First Diffusion. We show that SFD improves the VAE reconstruction-generation trade-off, accelerates DiT training convergence, and scales better with model size.

\paragraph{Towards better reconstruction performance.}
The tension between reconstruction fidelity and generation difficulty has long been a fundamental challenge in latent diffusion modeling \citep{esser2024scaling,yu2025representation}. It is essentially a dilemma: when the latent space preserves more information from the original image, the diffusion model must account for a richer and more complex distribution, demanding larger model capacity and slower convergence. Conversely, when the latent space is more heavily compressed (less information contained), modeling becomes easier because the diffusion model faces a easier and smoother distribution. This is precisely why methods that operate on pure semantic representations extracted via some visual foundation models~\citep{oquab2024dinov2} can converge rapidly \citep{zheng2025diffusion,tong2026scaling,svg2025,shi2025svg}. However, the forward propagation of a visual foundation model follows a Markov process in which semantic abstraction is inevitably accompanied by significant information loss, resulting in poor signal fidelity when reconstructing back to pixel space. Purely modeling on such representations would constrain the upper bound of consistency in fine-grained editing tasks and degrade the rendering of small text.

SFD offers a principled resolution to this dilemma. By introducing a semantic latent with inherently small capacity that discards information-dense but semantically redundant texture details, yet retains rich high-level semantic content, SFD provides cleaner guidance for texture generation. This can be viewed as supplying a stronger condition when modeling the texture latent: with richer conditioning, the distribution the model must capture becomes narrower, thereby simplifying generation. In this sense, SFD serves as a natural bridge between reconstruction and generation, rather than sacrificing one for the other.


Therefore, we can aggressively fine-tune the texture VAE toward reconstruction performance, as described in Section~\ref{subsec:architecture}. As illustrated in Table~\ref{tab:kodak-vae-reconstruction}, our fine-tuned FLUX.2 VAE improves PSNR from 33.18 to 36.40 on Kodak. Table~\ref{tab:omnidoc-vae-reconstruction} further shows that on the text-rich OmniDoc-TokenBench introduced by Qwen-Image-VAE-2.0 \citep{zhang2026qwenimagevae}, our VAE achieves the best PSNR, LPIPS, FID, and NED among selected baselines without any specialized training on text-rich data.

\paragraph{SFD accelerates training convergence.}
To study how SFD accelerates training convergence and improves the reconstruction–generation trade-off, we compare three configurations trained on 50M internal image-text samples (image resolution $256\times256$, learning rate $1\times10^{-4}$, global batch size 512, 32$\times$A800 GPUs): (i) fine-tuned FLUX.2 VAE without SFD, (ii) vanilla FLUX.2 VAE without SFD, and (iii) fine-tuned FLUX.2 VAE with SFD (ours). Evaluation is conducted on GenEval \citep{ghosh2023geneval} and DPG \citep{hu2024ella}.

\begin{figure}[t]
  \centering
  \includegraphics[width=\linewidth]{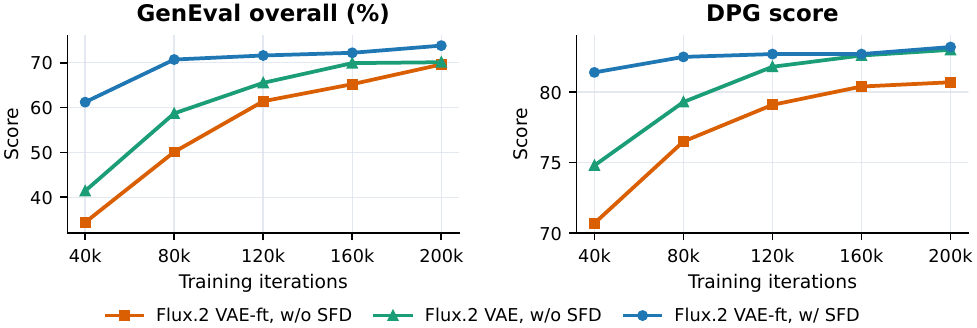}
  \caption{Training convergence comparison under the same 50M internal data setting. SFD converges faster than non-SFD baselines and maintains better final performance on GenEval and DPG.}
  \label{fig:sfd-convergence}
\end{figure}

As shown in Figure~\ref{fig:sfd-convergence}, SFD converges substantially faster and maintains a consistent lead on GenEval and DPG throughout training. This confirms that semantic guidance yields the same learnability benefit otherwise obtained by relaxing the VAE's latent distribution, without incurring the associated reconstruction penalty.

\begin{table}[t]
  \centering
  \caption{VAE reconstruction quality on Kodak \citep{kodakphotocd}. 
  Metrics include PSNR, SSIM \citep{wang2004ssim}, and LPIPS \citep{zhang2018perceptual}. Our fine-tuned FLUX.2 VAE substantially improves reconstruction fidelity.
  }
  \label{tab:kodak-vae-reconstruction}
  \sefitablefont
  \begin{tabular}{lccc}
    \toprule
    VAE & PSNR$\uparrow$ & SSIM$\uparrow$ & LPIPS$\downarrow$ \\
    \midrule
    SD1.5 & 26.66 & 0.7294 & 0.1452 \\
    FLUX.1 & 32.37 & 0.9063 & 0.0554 \\
    FLUX.2 & 33.18 & 0.9194 & 0.0442 \\
    \textbf{FLUX.2-finetuned (ours)} & \textbf{36.40} & \textbf{0.9565} & \textbf{0.0235} \\
    \bottomrule
  \end{tabular}
\end{table}

\begin{table}[t]
  \centering
  \caption{Selected VAE comparison on OmniDoc-TokenBench ($\sim$3K text-rich images, $256{\times}256$). Baseline results are selected from the Qwen-Image-VAE-2.0 evaluation. FID follows the standard Fréchet Inception Distance metric \citep{heusel2017gans}. Our result is measured on 3042 samples.}
  \label{tab:omnidoc-vae-reconstruction}
  \sefitablefont
  \begin{tabular}{lccccc}
    \toprule
    Model & PSNR$\uparrow$ & SSIM$\uparrow$ & LPIPS$\downarrow$ & FID$\downarrow$ & NED$\uparrow$ \\
    \midrule
    RAE-DINOv2-B & 14.32 & 0.3261 & 0.2290 & 18.21 & 0.0392 \\
    VAVAE & 17.50 & 0.6905 & 0.0974 & 4.45 & 0.3488 \\
    HunyuanImage-3.0 & 22.66 & 0.8672 & 0.0650 & 3.49 & 0.7753 \\
    Wan2.2 & 21.67 & 0.8577 & 0.0525 & 3.05 & 0.8310 \\
    FLUX.2 & 27.72 & 0.9544 & 0.0216 & 0.73 & 0.9535 \\
    Qwen-Image-VAE-2.0-f16c128 & 30.45 & 0.9706 & 0.0167 & 0.79 & 0.9617 \\
    \textbf{FLUX.2-finetuned (ours)} & \textbf{30.91} & \textbf{0.9718} & \textbf{0.0133} & \textbf{0.46} & \textbf{0.9648} \\
    \bottomrule
  \end{tabular}
\end{table}

\begin{figure}[t]
  \centering
  \includegraphics[width=0.78\linewidth]{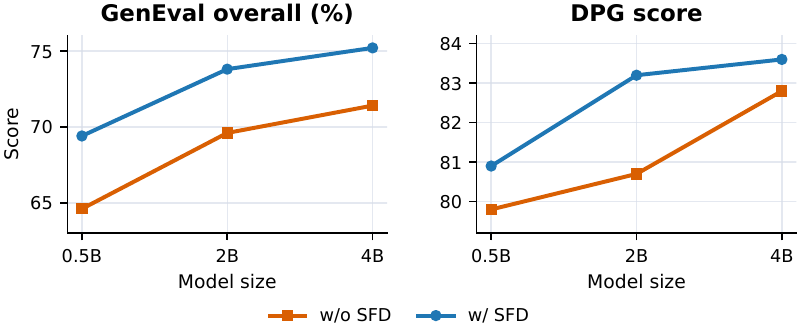}
  \caption{Model scaling comparison with and without SFD. SFD consistently outperforms the non-SFD baseline at the same model size, and a smaller SFD model can compete with or outperform a larger non-SFD model.}
  \label{fig:sfd-model-scaling}
\end{figure}

\begin{table}[t]
  \centering
  \caption{Training schedule of the DiT backbone. The pre-training stage follows a resolution curriculum, and each stage is initialized from the previous checkpoint.}
  \label{tab:training-schedule}
  \sefitablefont
  \begin{tabular*}{\linewidth}{@{\extracolsep{\fill}}llcccccc@{}}
    \toprule
    Stage & Data & Resolution & Batch size & $\Delta t$ & $\beta$ & Iterations & LR \\
    \midrule
    \multirow{4}{*}{Pre-training}
      & \multirow{4}{*}{450M}
      & 256px & 768 & 0.2 & 2 & 250K & $1{\times}10^{-4}$ \\
      &      & 512px & 768 & 0.2 & 2 & 300K & $5{\times}10^{-5}$ \\
      &      & 768px & 384 & 0.1 & 2 & 100K & $2{\times}10^{-5}$ \\
      &      & 1024px & 192 & 0.1 & 2 & 100K & $2{\times}10^{-5}$ \\
    \midrule
    Continual training & 9M & 1024px & 192 & 0.1 & 1 & 180K & $1{\times}10^{-5}$ \\
    Supervised fine-tuning & 650K & 1024px & 192 & 0.1 & 1 & 10K & $1{\times}10^{-5}$ \\
    \bottomrule
  \end{tabular*}
\end{table}

\paragraph{SFD with model scaling.}
We further investigate whether the advantage of semantic guidance persists at larger model sizes. Figure~\ref{fig:sfd-model-scaling} compares performance with and without SFD at 0.5B, 2B, and 4B model scales.
The advantage holds across all scales. Notably, the 2B SFD model even outperforms the 4B model without SFD by a significant margin, indicating that semantic-first modeling improves parameter efficiency and enables smaller SFD models to match or exceed larger counterparts.

\label{subsec:studies}


\section{Training}
\label{sec:training}

We train the DiT backbone in the composite semantic-texture latent space while keeping all encoders (Semantic VAE, Texture VAE, Qwen3-VL text encoder) frozen. Training proceeds in three stages: pre-training, continual high-resolution training, and supervised fine-tuning. We provide three model variants at 1B, 2B, and 5B parameters, all following the same training pipeline.


\subsection{Pre-training}
\label{subsec:training-pretraining}

Models at all scales follow the same curriculum learning schedule: 256px $\rightarrow$ 512px $\rightarrow$ 768px $\rightarrow$ 1024px, with each stage initialized from the previous checkpoint. The complete training schedule is summarized in Table~\ref{tab:training-schedule}. All stages use the full 450M recaptioned corpus mixed with synthetic text-rendered data (Section~\ref{subsec:data-pretraining}). Free-aspect-ratio training is enabled at every stage via predefined aspect-ratio buckets: 16:9, 4:3, 3:2, 1:1, 3:4, 2:3, and 9:16. Exponential moving average (EMA) with a decay rate of 0.9999 is applied throughout pre-training and all subsequent training stages.


\subsection{Continual Training}
\label{subsec:training-continual}
After the pre-training stage, the model acquires the basic capability to generate diverse elements. To further boost generation quality and instruction following capability, we continue training at 1024px on a more curated dataset. 
Starting from the 1024px pre-training checkpoint, we continue training on a curated mixture of high-quality recaptioned images. The learning rate is reduced to $1\times10^{-5}$. 

\subsection{Supervised Fine-Tuning}
\label{subsec:training-sft}

SFT narrows the output distribution toward high-quality, instruction-following generation. We train at 1024px on a score-refined dataset emphasizing hard prompts, strong aesthetics, accurate text rendering, and bilingual dense captions (with short captions and tags retained at lower weight for prompt-granularity robustness). The maximum text encoder context length is increased from 512 to 1024 to accommodate longer prompts.


\subsection{Few-Step Distillation}
\label{subsec:training-distillation}
To reduce inference cost, we distill the SFT model into a 4-step generator using DMD2 \citep{yin2024improved}. We focus here on our adaptation to SFD's dual-stream architecture.
\paragraph{Dual-stream schedule preservation.}
The key challenge is that directly compressing the teacher's 50-step trajectory into four student steps would remove the semantic-first offset. We therefore keep the same offset rule during distillation: timesteps for semantics consistently lead texture by $\Delta t=0.1$ for the student's 4-step generation. This preserves the three-phase schedule in Sec. \ref{sec:method} while reducing the number of sampling steps.
\paragraph{Training details.}
The teacher is the frozen SFT model sampled with its full multi-step schedule, and both the student and fake-score network are initialized from the teacher. The distillation process is trained at 1024px with the combined DMD matching loss, fake-score regression loss, and feature-space adversarial loss.

\subsection{RL Post-training}
\label{subsec:rl-post-training}

We apply DiffusionNFT~\citep{zheng2025diffusionnft} as an RL post-training stage to sharpen prompt following, visual quality, artifact suppression, and text rendering. The objective and loss formulation are detailed in Appendix~\ref{app:diffusionnft-objective}; here we describe how it is adapted to our dual-latent setting and the engineering choices that make online RL practical.

\paragraph{Adaptation to dual-latent space.}
DiffusionNFT operates on a clean-sample target reconstructed from the generated image. In our case, this target is the composite latent $z_{\mathrm{comp}} = \operatorname{concat}(z_{\mathrm{semantic}}, z_{\mathrm{texture}})$, obtained by re-encoding generated samples through both VAE branches. The asynchronous denoising schedule in Sec.~\ref{subsec:modeling} is kept intact; the RL objective only reshapes the reward-to-loss mapping without altering the generation dynamics.

\paragraph{Online iteration loop.}
Each iteration $i$ proceeds as:

\begin{equation}
\pi_i \xrightarrow{\mathrm{generate}} \mathrm{score} \rightarrow \mathrm{filter} \rightarrow \mathrm{train} \rightarrow \pi_{i+1}.
\end{equation}

The policy $\pi_i$ generates $M=12$ candidates for each of $K=400$ prompt groups, yielding 4,800 images per iteration. The full batch is scored before any gradient update; the generation checkpoint serves as the old-policy anchor in the DiffusionNFT loss.

\paragraph{Reward filtering and sample selection.}
Prompt groups with low reward dispersion, measured by reward standard deviation or range, are discarded because they carry little preference signal. Among retained groups, we apply top-bottom selection: high-reward samples provide positive gradients, while low-reward samples serve as implicit negatives.

\paragraph{Prompt and reward design.}
We treat online RL as an environment-feedback loop: the environment is the prompt distribution, and the feedback is a tagged scalar reward model. Prompts are selected for consistent evaluability rather than sampled uniformly. Each prompt carries capability tags, such as spatial composition, text rendering, and artifact control, and rewards are scored along the relevant dimensions. This tag-aware design keeps feedback capability-specific, reducing the risk that visually appealing but semantically incorrect samples are reinforced.

%% file: chapters/04_experiments.tex

%% file: chapters/05_performance_evaluation.tex

\section{Performance Evaluation}
\label{sec:performance-evaluation}

\begin{table}[t]
  \centering
  \caption{GenEval benchmark results.}
  \label{tab:geneval-performance}
  \sefitablefont
  \begin{tabular}{lccccccc}
    \toprule
    Model & Single Obj. & Two Obj. & Counting & Colors & Position & Attr. Binding & Overall$\uparrow$ \\
    \midrule
    \rowcolor{SeFiRowBlue}
    \textbf{SeFi-Image-5B} & 1.00 & 0.92 & 0.87 & 0.90 & 0.84 & 0.76 & 0.88 \\
    \rowcolor{SeFiRowBlue}
    \textbf{SeFi-Image-2B} & 0.99 & 0.93 & 0.84 & 0.91 & 0.78 & 0.78 & 0.87 \\
    \rowcolor{SeFiRowBlue}
    \textbf{SeFi-Image-1B} & 0.99 & 0.91 & 0.83 & 0.92 & 0.82 & 0.75 & 0.87 \\
    Qwen-Image & 0.99 & 0.92 & 0.89 & 0.88 & 0.76 & 0.77 & 0.87 \\
    FLUX.2-Klein-9B & 1.00 & 0.96 & 0.85 & 0.91 & 0.70 & 0.68 & 0.85 \\
    Z-Image & 1.00 & 0.94 & 0.78 & 0.93 & 0.62 & 0.77 & 0.84 \\
    \bottomrule
  \end{tabular}
\end{table}

\begin{table}[t]
  \centering
  \caption{DPG-Bench results.}
  \label{tab:dpg-performance}
  \sefitablefont
  \begin{tabular}{lcccccc}
    \toprule
    Model & Global & Entity & Attribute & Relation & Other & Overall$\uparrow$ \\
    \midrule
    Qwen-Image & 91.32 & 91.56 & 92.02 & 94.31 & 92.73 & 88.32 \\
    Z-Image & 93.39 & 91.22 & 93.16 & 92.22 & 91.52 & 88.14 \\
    JoyAI-Image & -- & -- & -- & -- & -- & 88.05 \\
    \rowcolor{SeFiRowBlue}
    \textbf{SeFi-Image-5B} & 88.24 & 92.62 & 91.49 & 93.57 & 91.55 & 87.27 \\
    \rowcolor{SeFiRowBlue}
    \textbf{SeFi-Image-2B} & 89.44 & 91.81 & 92.02 & 93.04 & 92.13 & 87.31 \\
    \rowcolor{SeFiRowBlue}
    \textbf{SeFi-Image-1B} & 91.19 & 93.16 & 91.59 & 92.13 & 86.66 & 87.17 \\
    Z-Image-Turbo & 91.29 & 89.59 & 90.14 & 92.16 & 88.68 & 84.86 \\
    \bottomrule
  \end{tabular}
\end{table}

We evaluate \method on prompt following, compositional reasoning, long-text rendering, visual text generation, and bilingual instruction generation. We report results for all three model variants (1B, 2B, and 5B) to study how performance scales under the semantic-first paradigm. Results are compared against strong open baselines including Qwen-Image \citep{wu2025qwenimage}, Z-Image \citep{zimage2025}, FLUX.2-Klein-9B \citep{flux2025frontier}, and JoyAI-Image \citep{song2026awaking}.

\paragraph{Prompt following and compositional reasoning.}
On GenEval \citep{ghosh2023geneval} (Table~\ref{tab:geneval-performance}), SeFi-Image-5B
achieves an overall score of 0.88, while the 1B and 2B variants both reach 0.87,
matching Qwen-Image and surpassing FLUX.2-Klein-9B (0.85) and Z-Image (0.84).
Notably, even our smallest 1B model
ties with the much larger Qwen-Image, suggesting that semantic-first modeling
provides strong compositional reasoning capabilities even at small scale. The
sub-metric breakdown reveals that \method is especially competitive on spatial
understanding (Position) and color fidelity, while Counting remains the primary
gap relative to Qwen-Image.

On DPG-Bench \citep{hu2024ella} (Table~\ref{tab:dpg-performance}), SeFi-Image-5B scores 87.27 overall, slightly below Qwen-Image (88.32) and Z-Image (88.14). The 2B and 1B variants remain within one point of the 5B model, showing comparable compositional performance across model scales.

\paragraph{Long-text rendering and visual text generation.}
LongTextBench \citep{geng2025xomni} (Table~\ref{tab:longtextbench-performance}) evaluates the model's
ability to follow lengthy, detailed and text-rich prompts. SeFi-Image-5B achieves the highest
average score (0.9780) among all models, surpassing JoyAI-Image (0.9630) and
Qwen-Image-2512 (0.9604), with balanced performance across English and Chinese.
This indicates that the semantic branch provides a strong structural scaffold
that helps the model organize complex, information-dense prompts. However, the
1B and 2B variants show a noticeable drop (0.85--0.87), revealing that long-text
comprehension still benefits substantially from increased model capacity. 
Even so, our 1B and 2B models already outperform FLUX.2-Klein-9B. Another interesting observation is that the 1B model is more balanced across English and Chinese, while the 2B model improves on English but drops on Chinese. This non-monotonic pattern may reflect data-distribution bias or training variance toward English-language performance, rather than a simple capacity-driven scaling trend.

\begin{table}[t]
  \centering
  \sefitablefont

  \begin{minipage}[t]{0.48\textwidth}
    \centering
    \caption{LongTextBench results.}
    \label{tab:longtextbench-performance}
    \begin{tabular}{lccc}
      \toprule
      Model & EN$\uparrow$ & ZH$\uparrow$ & Avg$\uparrow$ \\
      \midrule
      \rowcolor{SeFiRowBlue}
      \textbf{SeFi-Image-5B} & 0.978 & 0.978 & 0.978 \\
      JoyAI-Image & 0.963 & 0.963 & 0.963 \\
      Qwen-Image-2512 & 0.956 & 0.965 & 0.960 \\
      Qwen-Image & 0.943 & 0.946 & 0.945 \\
      Z-Image & 0.935 & 0.936 & 0.936 \\
      \rowcolor{SeFiRowBlue}
      \textbf{SeFi-Image-1B} & 0.854 & 0.856 & 0.855 \\
      \rowcolor{SeFiRowBlue}
      \textbf{SeFi-Image-2B} & 0.870 & 0.824 & 0.847 \\
      FLUX.2-Klein-9B & 0.864 & 0.218 & 0.541 \\
      \bottomrule
    \end{tabular}
  \end{minipage}
  \hfill
  \begin{minipage}[t]{0.48\textwidth}
    \centering
    \caption{CVTG-2k results.}
    \label{tab:cvtg-performance}
    \begin{tabular}{lccc}
      \toprule
      Model & NED$\uparrow$ & CLIPScore$\uparrow$ & Word Acc.$\uparrow$ \\
      \midrule
      \rowcolor{SeFiRowBlue}
      \textbf{SeFi-Image-5B} & 0.943 & 0.816 & 0.895 \\
      JoyAI-Image & 0.937 & 0.799 & 0.874 \\
      Z-Image & 0.937 & 0.797 & 0.867 \\
      Z-Image-Turbo & 0.928 & 0.805 & 0.859 \\
      Qwen-Image & 0.912 & 0.802 & 0.829 \\
      \rowcolor{SeFiRowBlue}
      \textbf{SeFi-Image-2B} & 0.896 & 0.816 & 0.773 \\
      \rowcolor{SeFiRowBlue}
      \textbf{SeFi-Image-1B} & 0.862 & 0.810 & 0.718 \\
      \bottomrule
    \end{tabular}
  \end{minipage}
\end{table}

On CVTG-2K \citep{tai2025complexvtg} (Table~\ref{tab:cvtg-performance}), which specifically measures
character-level text rendering accuracy, SeFi-Image-5B achieves 0.8947 Word
Accuracy and 0.9434 NED, improving over the strongest baselines on these two
text-rendering metrics. The smaller
variants show progressive degradation (2B: 0.77, 1B: 0.72 Word Acc.).
Meanwhile, we observe that all variants of \method with different model sizes achieve high CLIPScore \citep{hessel2021clipscore}, indicating that semantic guidance effectively improves alignment between images and text.

\paragraph{Bilingual instruction generation.}
The OneIG benchmarks \citep{chang2025oneig} (Tables~\ref{tab:oneig-zh-performance}
and~\ref{tab:oneig-en-performance}) assess overall generation quality under
complex bilingual instructions, encompassing alignment, text rendering,
reasoning, style, and diversity. On OneIG-EN
(Table~\ref{tab:oneig-en-performance}), SeFi-Image-5B achieves the best overall
score (0.5606), outperforming Z-Image (0.5460) and Qwen-Image (0.5390). On OneIG-ZH
(Table~\ref{tab:oneig-zh-performance}), it reaches 0.5379 overall, slightly above
Z-Image while using limited data and training cost.

\begin{table}[t]
  \centering
  \caption{OneIG benchmark results.}
  \label{tab:oneig-performance}
  \sefitablefont

  \begin{subtable}[t]{\textwidth}
    \centering
    \caption{OneIG-ZH}
    \label{tab:oneig-zh-performance}
    \begin{tabular}{lcccccc}
      \toprule
      Model & Alignment & Text & Reasoning & Style & Diversity & Overall$\uparrow$ \\
      \midrule
      Qwen-Image & 0.8250 & 0.9630 & 0.2670 & 0.4050 & 0.2790 & 0.5480 \\
      Z-Image & 0.7930 & 0.9880 & 0.2660 & 0.3860 & 0.2430 & 0.5350 \\
      \rowcolor{SeFiRowBlue}
      \textbf{SeFi-Image-5B} & 0.8063 & 0.9716 & 0.2672 & 0.4307 & 0.2137 & 0.5379 \\
      JoyAI-Image & -- & -- & -- & -- & -- & 0.5210 \\
      \rowcolor{SeFiRowBlue}
      \textbf{SeFi-Image-1B} & 0.7920 & 0.9022 & 0.2394 & 0.3934 & 0.2206 & 0.5095 \\
      \rowcolor{SeFiRowBlue}
      \textbf{SeFi-Image-2B} & 0.7974 & 0.8120 & 0.2476 & 0.4043 & 0.2280 & 0.4979 \\
      FLUX.2-Klein-9B & 0.8201 & 0.4920 & 0.2599 & 0.4166 & 0.1625 & 0.4302 \\
      FLUX.2-Klein-9B (Reimp) & 0.7849 & 0.3450 & 0.2608 & 0.4219 & 0.1612 & 0.3948 \\
      \bottomrule
    \end{tabular}
  \end{subtable}

  \vspace{0.8em}

  \begin{subtable}[t]{\textwidth}
    \centering
    \caption{OneIG-EN}
    \label{tab:oneig-en-performance}
    \begin{tabular}{lcccccc}
      \toprule
      Model & Alignment & Text & Reasoning & Style & Diversity & Overall$\uparrow$ \\
      \midrule
      \rowcolor{SeFiRowBlue}
      \textbf{SeFi-Image-5B} & 0.8648 & 0.9576 & 0.3240 & 0.4502 & 0.2065 & 0.5606 \\
      Z-Image & 0.8810 & 0.9870 & 0.2800 & 0.3870 & 0.1940 & 0.5460 \\
      JoyAI-Image & -- & -- & -- & -- & -- & 0.5420 \\
      Qwen-Image & 0.8820 & 0.8910 & 0.3060 & 0.4180 & 0.1970 & 0.5390 \\
      \rowcolor{SeFiRowBlue}
      \textbf{SeFi-Image-2B} & 0.8664 & 0.9017 & 0.2794 & 0.4209 & 0.2150 & 0.5367 \\
      FLUX.2-Klein-9B & 0.8871 & 0.8657 & 0.3117 & 0.4417 & 0.1560 & 0.5324 \\
      \rowcolor{SeFiRowBlue}
      \textbf{SeFi-Image-1B} & 0.8594 & 0.8839 & 0.2706 & 0.3975 & 0.2121 & 0.5247 \\
      FLUX.2-Klein-9B (Reimp) & 0.8684 & 0.8385 & 0.3130 & 0.4449 & 0.1562 & 0.5241 \\
      \bottomrule
    \end{tabular}
  \end{subtable}
\end{table}

\paragraph{Summary.}
Taken together, these results demonstrate that \method achieves competitive or state-of-the-art performance across diverse evaluation axes with significantly less training compute (125K A800 GPU hours). The 5B model is the strongest overall, while the 1B and 2B variants remain surprisingly competitive on several tasks, where semantic-first guidance effectively compensates for reduced model capacity. Evaluation of the few-step turbo variants is provided in Appendix~\ref{app:distillation-details}.

%% file: chapters/06_visualization.tex

\section{Visualization}
\label{sec:visualization}

\begin{figure}[H]
  \centering
  \vspace*{0.14\paperheight}
  \makebox[\linewidth][c]{%
    \includegraphics[width=0.94\paperwidth,height=0.45\paperheight,keepaspectratio]{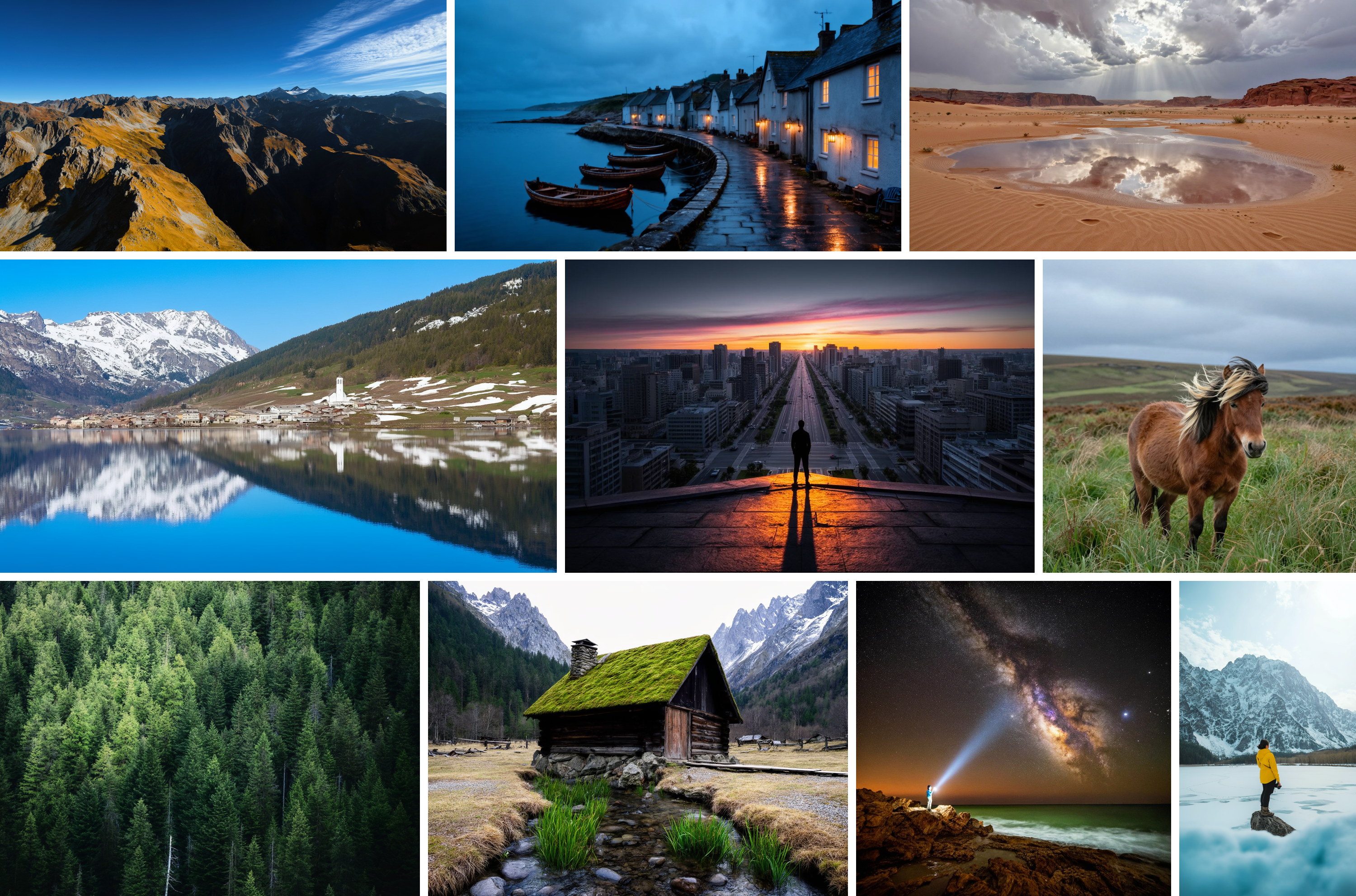}%
  }
  \caption{Natural scene examples generated by \method across multiple aspect ratios.}
  \label{fig:visualization-scene-ratio}
\end{figure}

\begin{figure}[p]
  \centering
  \makebox[\linewidth][c]{%
    \includegraphics[width=0.96\paperwidth,height=0.84\paperheight,keepaspectratio]{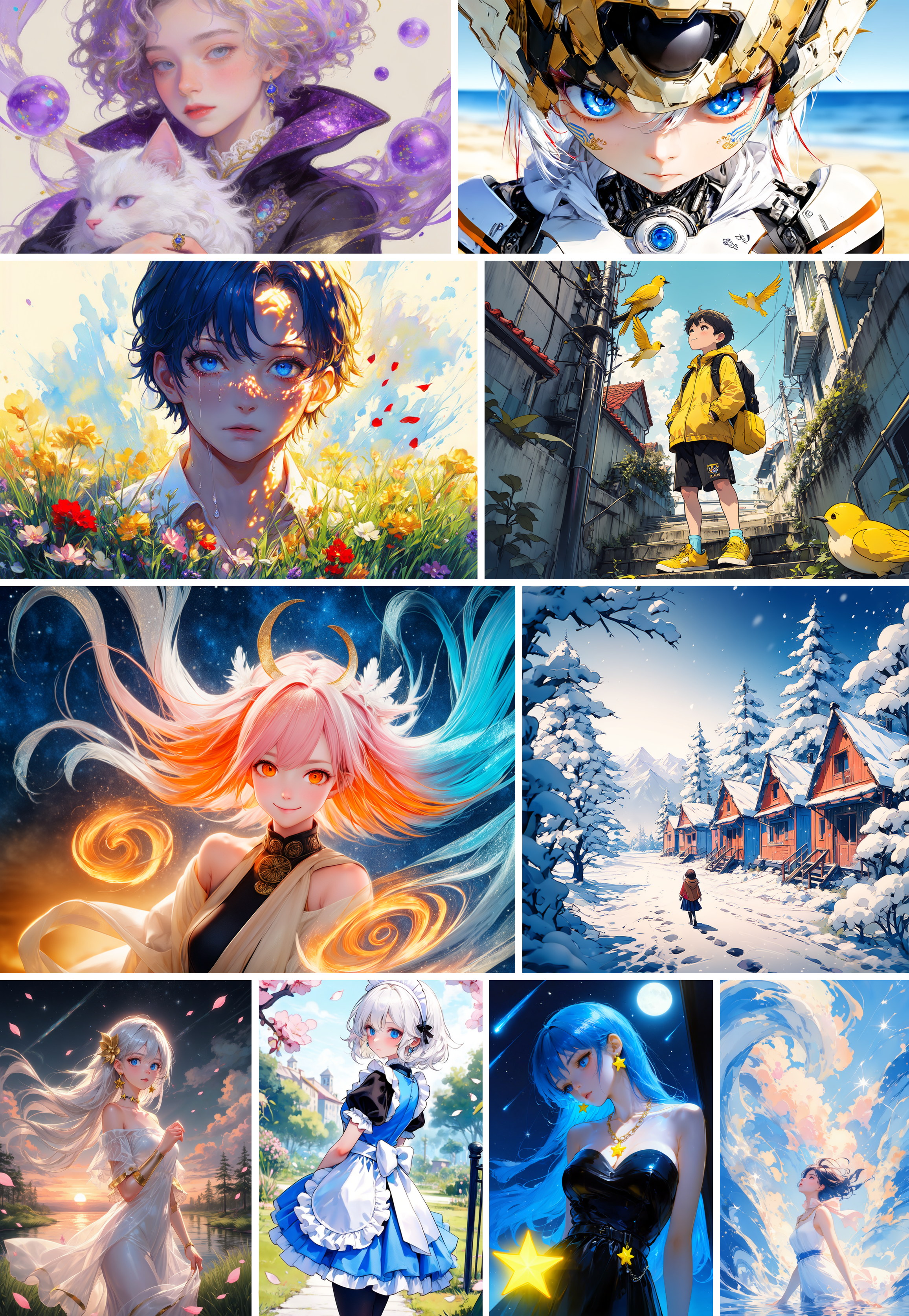}%
  }
  \caption{Anime-style examples generated by \method across multiple aspect ratios.}
  \label{fig:visualization-anime-ratio}
\end{figure}

\begin{figure}[p]
  \centering
  \makebox[\linewidth][c]{%
    \includegraphics[width=0.96\paperwidth,height=0.84\paperheight,keepaspectratio]{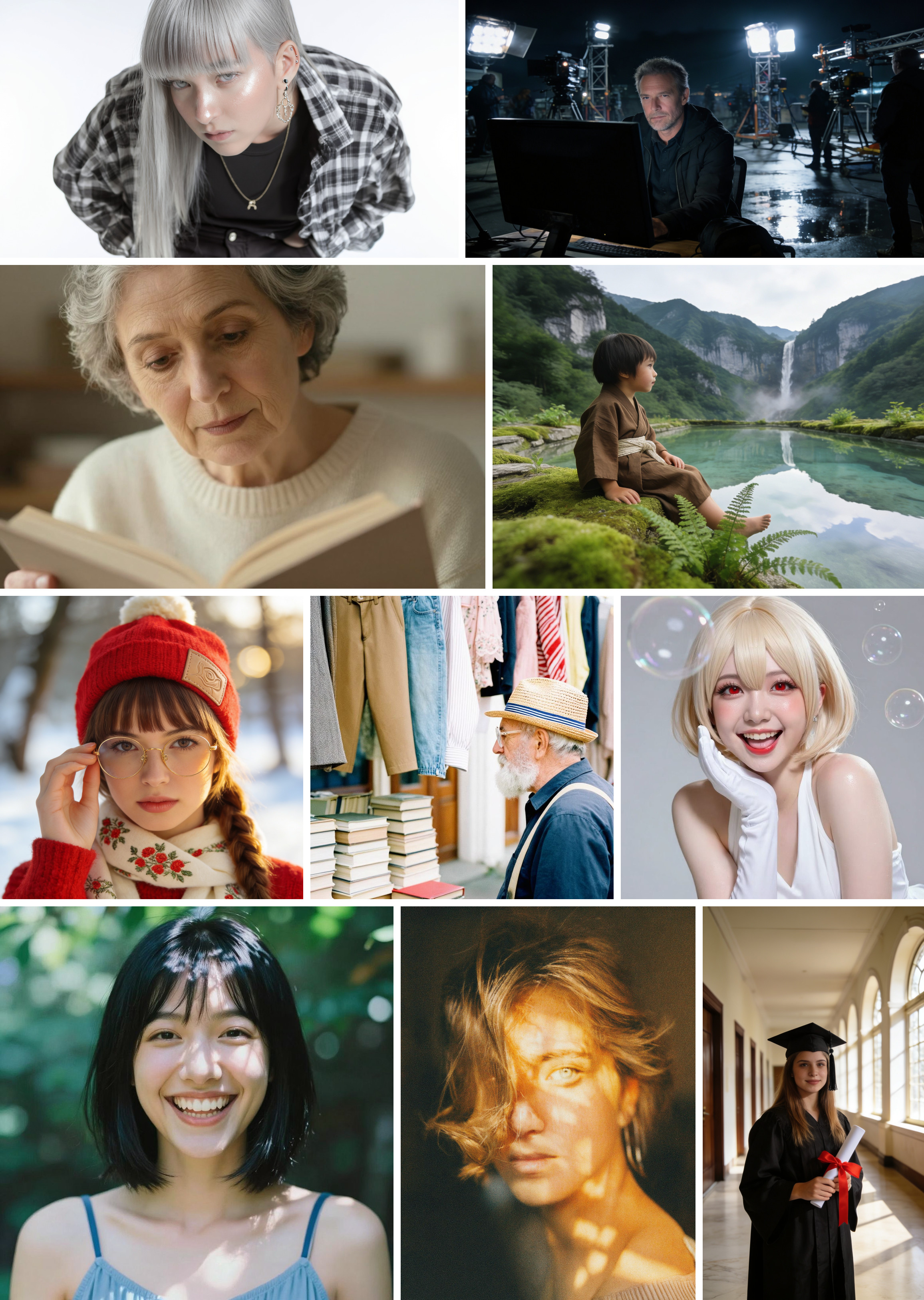}%
  }
  \caption{Portrait examples generated by \method across multiple aspect ratios.}
  \label{fig:visualization-portrait-ratio}
\end{figure}

\begin{figure}[p]
  \centering
  \makebox[\linewidth][c]{%
    \includegraphics[width=0.96\paperwidth,height=0.84\paperheight,keepaspectratio]{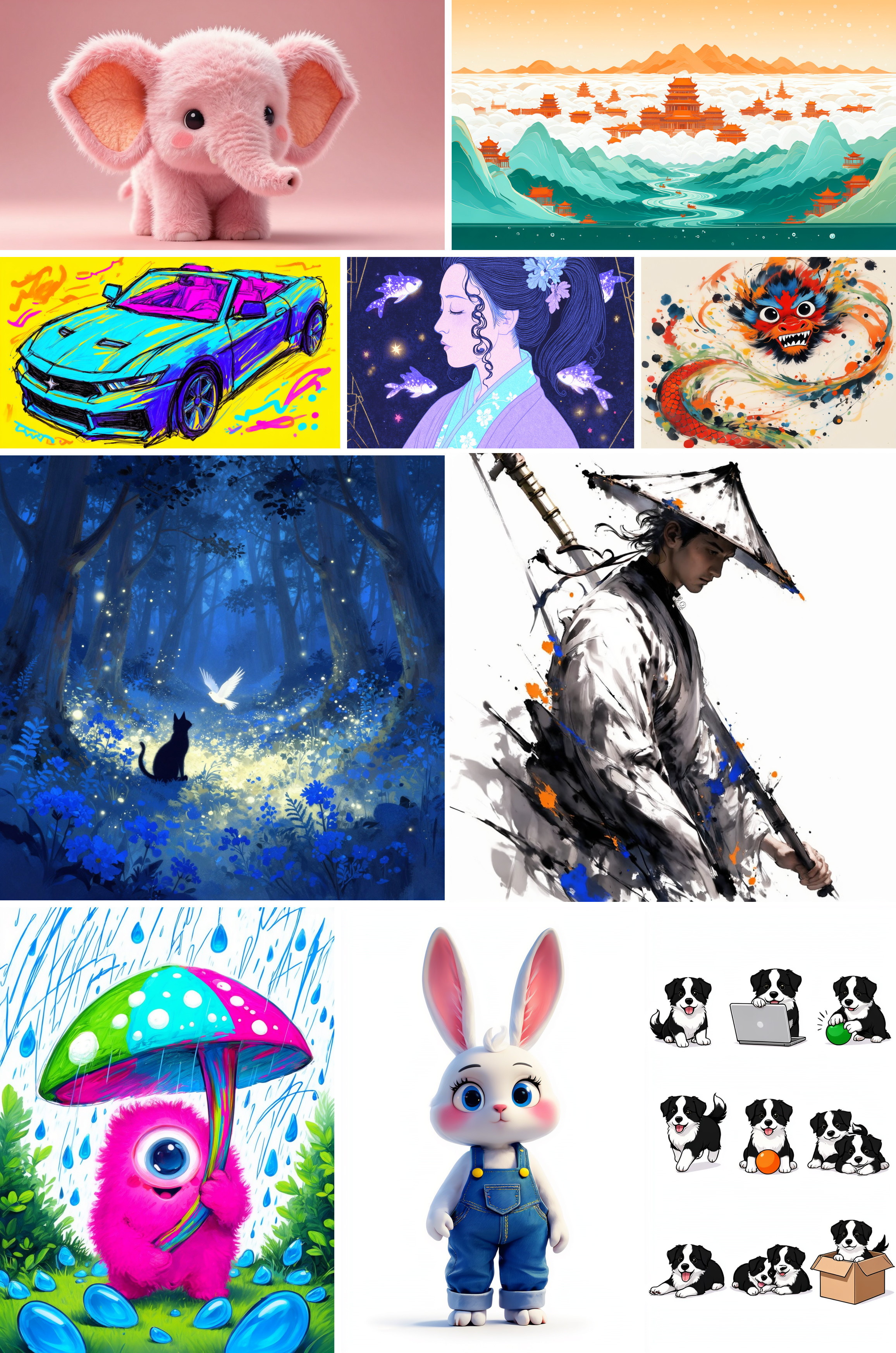}%
  }
  \caption{Stylized generation examples produced by \method across multiple aspect ratios.}
  \label{fig:visualization-style-ratio}
\end{figure}

\begin{figure}[p]
  \centering
  \makebox[\linewidth][c]{%
    \includegraphics[width=0.96\paperwidth,height=0.84\paperheight,keepaspectratio]{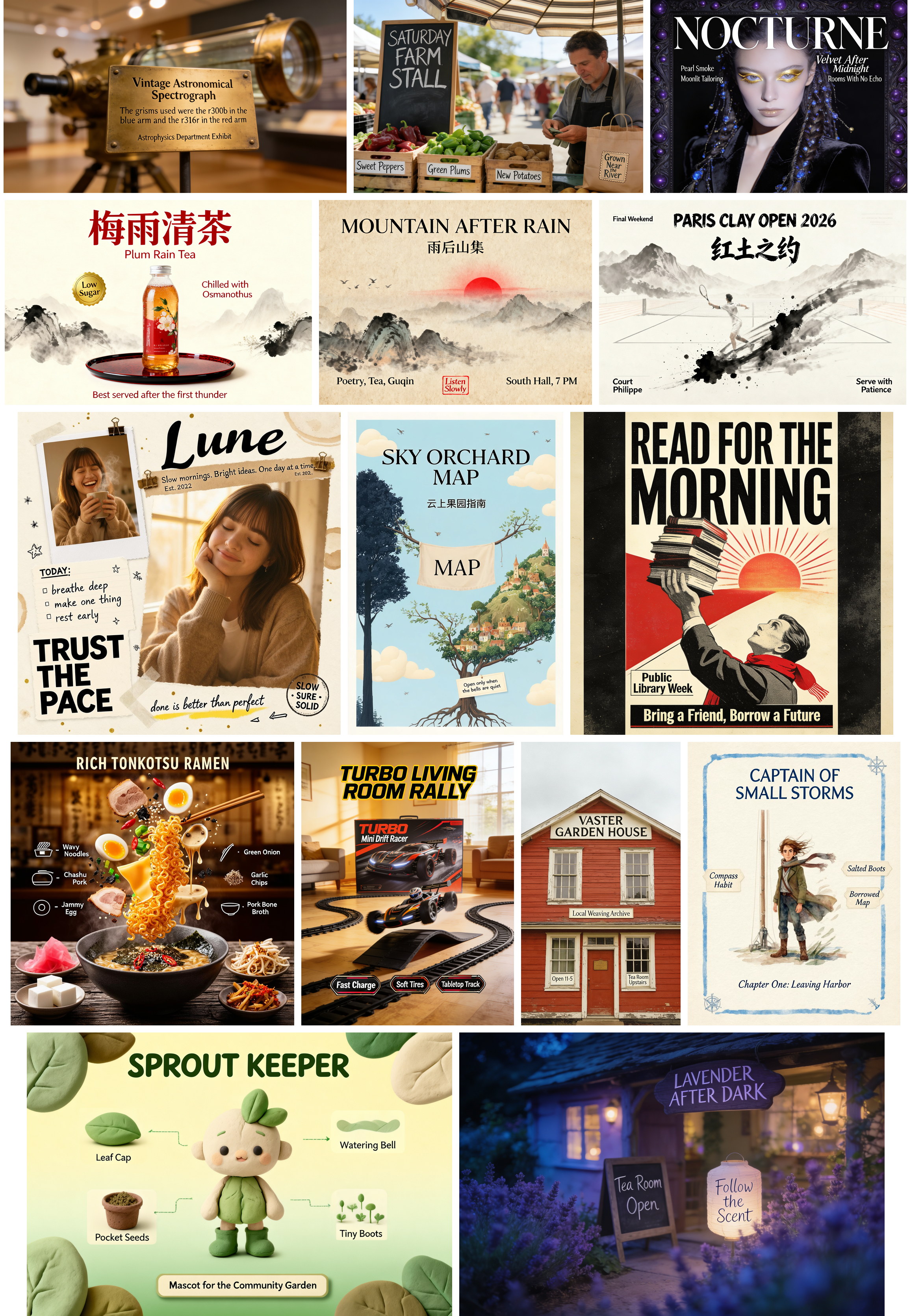}%
  }
  \caption{Text-rich examples generated by \method across multiple aspect ratios.}
  \label{fig:visualization-text-ratio}
\end{figure}

Figures~\ref{fig:visualization-scene-ratio}--\ref{fig:visualization-text-ratio}
show qualitative results from \method across natural scenes, anime-style images,
portraits, stylized generation, and text-rich layouts. Each canvas mixes square,
landscape, and portrait outputs to illustrate visual diversity and
free-aspect-ratio generation; the text-rich canvas further includes posters,
signs, labels, maps, and menu-like designs with readable rendered text.

\clearpage
\newgeometry{left=0.28in,right=0.28in,top=0.25in,bottom=0.50in}

\clearpage

\clearpage

\clearpage

\clearpage
\newgeometry{left=1.5in,right=1.5in,top=1in,bottom=1in}

%% file: chapters/05_limitations.tex

\section{Limitations and Future Work}
\label{sec:limitations}

\textbf{Insufficient scaling.} The primary limitation of SeFi-Image is insufficient scaling along the axes of model size, data, and compute. Although our 1B-to-5B model family has enabled us to explore the scaling behavior and potential of SFD for text-to-image generation, hardware constraints, specifically the use of NVIDIA A800 40G GPUs, limit our largest model to 5B parameters. Model scale is closely tied to several capabilities critical for text-to-image generation, including accurate text rendering, precise layout construction, and fine-grained prompt following. Based on the observed scaling trends, we expect that further increasing model capacity would yield additional gains. On the data side, both our pre-training and continual-training sets remain relatively limited in scale, and we have not yet explored highly refined data mixtures, leaving substantial room for improvement. In future work, we plan to scale SeFi-Image to larger model sizes, train on richer, higher-quality, and more broadly distributed data, and extend training under larger compute budgets.

\textbf{Insufficient training data exploration.} Our pre-training corpus is biased toward natural images, with relatively sparse coverage of aesthetic, artistic, design-oriented, screen-UI, and graphic-design content. This constrains both the quality ceiling and the stylistic diversity of the model. Moreover, we have not yet incorporated a more sophisticated agent-based synthetic data pipeline. Consequently, the model exhibits weaker performance on tasks requiring infographics, structured visual explanations, or complex graphic-design layouts.

\textbf{Multimodal generation unexplored.} A core motivation of SeFi-Image is to improve the reconstruction–generation trade-off, yet we have not validated this advantage on image editing tasks, where reconstruction fidelity and content consistency are even more critical. For instance, many editing scenarios require selective modification of certain regions or attributes while precisely preserving the rest of the image. Evaluating and extending SeFi-Image for such image-conditioned generation and editing tasks remains an important direction for future work.

\textbf{Improving the reconstruction and generation trade-off of video generation.} Finally, Semantic-First Diffusion (SFD) \citep{pan2025semantics} offers a promising direction for video generation. Existing video generators typically employ VAEs with aggressive compression ratios to encode video into latent spaces, resulting in considerable information loss. While increasing the latent channel capacity improves reconstruction quality, it also makes the latent space richer and more complex, rendering diffusion modeling and convergence more difficult. In semantic-first modeling, the semantic latent capacity is relatively fixed, suggesting a different trade-off: one can allocate greater capacity to texture latents to improve reconstruction while keeping semantic modeling stable. This may yield a more favorable reconstruction–generation balance for future video generation systems.

%% file: chapters/07_conclusion.tex

\section{Conclusion}
\label{sec:conclusion}

We presente SeFi-Image, a text-to-image foundation model built upon Semantic-First Diffusion, which decouples the denoising process into an asynchronous semantic-texture schedule to resolve the reconstruction–generation trade-off in latent diffusion models. By instantiating three model variants at 1B, 2B, and 5B parameters, we demonstrate that semantic-first modeling not only transfers effectively from small-scale class-conditional settings to large-scale, high-resolution text-to-image generation, but also yields significant training efficiency gains—our largest 5B model requires only 125K A800 GPU hours (approximately 10–20\% of Z-Image) while achieving competitive or superior performance. Scaling experiments further confirm that SFD consistently improves parameter efficiency without diminishing returns at larger model sizes, suggesting that structurally separating semantic layout from texture synthesis is a principled and resource-efficient paradigm for building text-to-image foundation models.

%% file: chapters/references.tex

\bibliographystyle{unsrtnat}
\bibliography{references}

%% file: chapters/appendix.tex

\newtcblisting{promptbox}{
  enhanced,
  breakable,
  listing only,
  colback=blue!2!white,
  colframe=blue!45!black,
  boxrule=0.6pt,
  arc=4pt,
  left=4pt,
  right=4pt,
  top=6pt,
  bottom=6pt,
  listing options={
    basicstyle=\ttfamily\tiny,
    breaklines=true,
    columns=fullflexible,
    keepspaces=true,
    showstringspaces=false
  }
}

\clearpage
\clearpage

\section{Additional Related Work}
\label{app:related-work}

\paragraph{Foundational text-to-image models.}
Recent text-to-image foundation models have made substantial progress in prompt
following, visual fidelity, text rendering, and image editing.
Qwen-Image improves complex text rendering and precise image editing with a
large-scale data pipeline and multi-task training recipe
\citep{wu2025qwenimage}. Seedream 3.0 and Seedream 4.0 further push
high-quality generation and multimodal image generation capabilities
\citep{gao2025seedream3,seedream2025seedream4}. Z-Image studies efficient
foundation-model training with a single-stream diffusion transformer and reports
strong performance under a relatively resource-friendly training setup
\citep{zimage2025}. JoyAI-Image moves toward a unified multimodal model that
supports visual understanding, text-to-image generation, and instruction-guided
editing through a shared multimodal interface \citep{song2026awaking}. These
systems show the rapid progress of modern text-to-image foundation models.
\method is complementary to this line of work: rather than only scaling model
size, data, or conditioning pipelines, it focuses on improving the
reconstruction--generation trade-off through semantic-first modeling.

\paragraph{VAEs for latent generation.}
The VAE is a central component of latent image generation because it defines the
space in which the generative model operates. Standard latent diffusion uses a
reconstruction-oriented VAE to compress images before diffusion
\citep{rombach2022high}. This design is efficient, but the latent space must
serve two different needs at the same time: it should preserve enough visual
detail for reconstruction, while also remaining easy for the generative model to
learn. Recent work studies this tension more directly. VA-VAE aligns the latent
space with visual foundation model features to improve semantic representation
\citep{yao2025reconstruction}, while RAE replaces the conventional VAE with
representations from pretrained visual encoders \citep{zheng2025diffusion}. In
contrast, \method keeps a high-fidelity texture VAE for image reconstruction and
uses a separate Semantic VAE to provide compact semantic guidance.

\paragraph{Semantic guidance for generation.}
Several recent methods show that adding semantic information from pretrained
visual encoders can make diffusion training easier. REPA aligns intermediate
diffusion features with pretrained visual representations
\citep{yu2025representation}. VA-VAE align the latent space to contain stronger semantic information
\citep{yao2025reconstruction}, and RAE force diffusion modeling on the semantic representation itself~\citep{zheng2025diffusion}. SFD further introduces an
asynchronous denoising schedule in which semantic latents are denoised ahead of
texture latents, so texture generation is guided by a cleaner semantic anchor
\citep{pan2025semantics}. These methods provide evidence that semantic
information can improve ImageNet class-conditional generation \citep{russakovsky2015imagenet}. \method studies
whether this idea remains useful in larger-scale, higher-resolution, and
open-ended text-to-image generation.

\section{Data Construction Details}
\label{app:data-construction}

\subsection{Pre-training Caption Prompt}
\label{app:pretrain-caption-prompt}

We annotate pre-training images with a VLM-based captioning pipeline. The
captioning prompt asks the annotator to generate bilingual captions grounded
only in visible image evidence. Each image receives both a concise
\texttt{short\_caption} and a more complete \texttt{dense\_caption}, and each
caption is provided in English and Chinese. The prompt used for pre-training
caption generation is shown below.

\begin{promptbox}
You are a professional image caption annotator for text-to-image data.
Generate faithful bilingual captions grounded only in visual evidence.

Core principles:
- Accuracy > richness. When unsure, omit or hedge instead of guessing.
- Describe only visually supported facts.
- English and Chinese captions must be semantically equivalent.
- short_caption must be a faithful subset of dense_caption: it may delete details but must not add, generalize, or change facts.
- If legible text exists in the image, transcribe it exactly in the original script; if only part is readable, include only the readable part and never guess missing text.
- Do not infer identity, job, age, nationality, intent, emotion, backstory, brand, material, function, time, or location unless clearly visible.
- If the image contains unusual, incorrect, or abnormal visual content, describe it explicitly instead of normalizing it.

Write bilingual captions for the image in English and Chinese.

Requirements:
1. Produce:
   - short_caption: concise and high-signal
   - dense_caption: complete, specific, and concise
2. Cover the main subjects, main action or state, main scene, key spatial
   relations, reliable counts, and clearly visible attributes when important.
3. short_caption must retain the core facts from dense_caption, especially the main subject, main action or state, main scene, and key count, OCR, or abnormal details when present.
4. Use cautious wording when details are uncertain because of blur, occlusion, crop, low resolution, overexposure, or partial visibility.
5. Use viewer perspective consistently for spatial relations such as left, right, top, bottom, front, and behind.
6. dense_caption should be objective and natural, without marketing language, storytelling, or aesthetic praise.
7. If legible text exists in the image, put quoted visible text in double quotes.

Output strict JSON only:
{
  "short_caption": {
    "en": "...",
    "zh": "..."
  },
  "dense_caption": {
    "en": "...",
    "zh": "..."
  }
}
\end{promptbox}

\subsection{SFT Metadata and Caption Prompts}
\label{app:sft-caption-prompt}

For supervised fine-tuning data, we use a two-stage VLM annotation workflow. The
first prompt extracts structured metadata, tags, safety attributes, quality
signals, OCR, and first-pass bilingual captions. The second prompt refines the
caption using the image and the first-pass metadata as context, with stricter
requirements for exhaustive visual description and OCR preservation.

\paragraph{Metadata extraction prompt.}
The prompt used for extracting structured metadata is shown below.

\begin{promptbox}
You are a strict image metadata annotator for a text-to-image SFT dataset.

Return exactly one JSON object. Do not use Markdown. Do not wrap the JSON in a code block. Do not add explanations outside the JSON.

Core caption and metadata principles:
- Accuracy is more important than richness. When unsure, omit or hedge instead of guessing.
- Describe only visually supported facts.
- Do not infer identity, job, age, nationality, intent, emotion, backstory, brand, material, function, time, or location unless clearly visible as text or visual evidence.
- If the image contains unusual, incorrect, broken, or abnormal visual content, describe it explicitly instead of normalizing it.
- Use viewer perspective consistently for left, right, top, bottom, foreground, background, front, and behind.

Use this category enum:
- landscape: natural or urban scenery where the place/scene is the subject
- portrait: one or more people as the main subject
- object: product-like or single-object focus that is not clearly e-commerce
- fauna_animals: animals are the main subject
- flora_plants: plants, flowers, trees, or botanical content is the main subject
- art: illustration, painting, anime, CG, concept art, or stylized artwork
- poster: graphic poster, advertisement poster, typography-heavy design poster
- slide: presentation slide, chart slide, teaching slide, infographic slide
- ui: app, website, software interface, dashboard, game UI, device screen UI
- e_commerce: product listing, catalog image, commercial product display
- others: use only when no category fits

Choose exactly one primary_category. Use secondary_categories for additional applicable categories, but do not include the primary category again.

Tag requirements:
- tags_en and tags_zh should include image style, image type, main subjects, scene elements, visual quality, layout, and likely use case when obvious.
- Prefer concise tags, 6 to 16 tags per language.

Safety requirements:
- Mark explicit nudity, pornographic framing, severe gore, or obvious violence.
- Mild fashion, swimwear, attractive portraits, or beautiful people are allowed when there is no explicit nudity and the image is not too sexually revealing.

Watermark requirement:
- has_intrusive_watermark should be true only when watermark/logo/signature is visually intrusive or seriously harms training value.

Text and OCR requirements:
- Text extraction is critical. If legible text exists, transcribe it exactly in the original script, character by character, into visible_text.
- visible_text[].text must contain only exact visible text or exact readable fragments. Never put generic descriptions such as "partial blurry sign" in the text field; put that in notes.
- visible_text is only for readable OCR. If a text-like region has no readable characters, do not create a visible_text item for it. Record it in unreadable_text_regions instead.
- If only part of a text region is readable, include only the readable fragment in visible_text and note the unreadable part in notes. Never guess missing characters.
- For each text region, include viewer-relative position and objective text style, such as "top center, large bold white sans-serif title" or "right side, red vertical Chinese calligraphy".
- For unreadable_text_regions, describe only the viewer-relative position, visual style, and why it is unreadable, such as blur, crop, occlusion, low resolution, or garbled rendering. Do not invent or paraphrase the text.
- Set has_non_chinese_english_text when visible text contains Japanese, Korean, Arabic, Cyrillic, or other non-Chinese/non-English writing.
- text_is_clear means all important visible text is readable.
- has_garbled_or_blurry_text means text appears corrupted, nonsensical, AI-garbled, or too blurry to read.

Quality requirements:
- is_blurry means the image itself is visibly blurry or out of focus, not just a shallow-depth-of-field artistic background.
- is_low_quality means low resolution, heavy compression artifacts, broken rendering, placeholder images, or otherwise poor training quality.
- has_corruption means broken files, missing image placeholders, severe glitches, blank/mostly blank images, or unreadable content.

Caption requirements:
- Produce four captions: short_en, short_zh, long_en, long_zh.
- English and Chinese captions must be semantically equivalent.
- short_en and short_zh must be faithful subsets of the long captions: concise, high-signal, and not adding, generalizing, or changing facts.
- If the image contains readable text, OCR completeness has priority over short-caption brevity. A short caption may become longer when needed to preserve all visible text.
- Every visible_text[].text string must appear verbatim in short_en, short_zh, long_en, and long_zh. Put each exact visible text string in double quotes in every caption.
- For any image with readable text, build short_en and short_zh with an explicit complete OCR list:
  - short_en should end with: Visible text: "text1"; "text2"; ...
  - short_zh should end with the Chinese visible-text prefix, followed by: "text1"; "text2"; ...
  Include every visible_text[].text item in this list. Never omit smaller UI labels, seals, signs, buttons, subtitles, or secondary text for brevity.
- Long captions should also contain every visible_text[].text item verbatim, preferably near the sentence that describes its location and visual style.
- Captions must not introduce extra transcribed or quoted text that is missing from visible_text. If a text-like region is unreadable, describe only that it is unreadable and where it appears; do not invent the text.
- long_en and long_zh must be objective, specific, and concise. Cover main subjects, main action/state, scene, key spatial relations, reliable counts, clearly visible attributes, image type/style, and abnormal details.
- Do not use marketing language, storytelling, aesthetic praise, or unsupported interpretation.
- If readable text exists, include all exact visible text in the captions, along with where it appears and what it looks like.
- If unreadable or garbled text exists, mention its position and that it is unreadable/garbled; do not invent the text.

Output JSON schema:
{{SCHEMA_JSON}}
\end{promptbox}

\paragraph{Detailed caption prompt.}
The prompt used for detailed caption generation is shown below.

\begin{promptbox}
You are a strict detailed image caption annotator for a text-to-image SFT dataset.

Return exactly one JSON object. Do not use Markdown. Do not wrap the JSON in a code block. Do not add explanations outside the JSON.

The first-pass category, tags, and OCR are provided below as context. Use the image as the source of truth. Use context to preserve recognized text and broad image type, but do not blindly trust context if the image contradicts it.

Context JSON:
{{CONTEXT_JSON}}

Core principles:
- Describe only visible content that exists in the image.
- Do not describe absent properties or negative checks.
- If the image has no visible text, do not mention text at all.
- Do not write data-quality report language inside captions.
- Do not infer identity, job, age, nationality, intent, relationship, emotion, backstory, brand, material, function, time, or location unless visibly supported.
- Do not summarize complex images. Prefer explicit, exhaustive visual description over compressed summaries, especially in long captions.
- The caption should objectively, completely, and faithfully match the visual evidence. Describe visible subjects, attributes, spatial relations, layout, occlusion, and interactions so the text can be matched back to the image.
- Be specific and concrete. Do not use vague phrases like "several people", "various objects", "multiple sections", or "etc." when a better count, approximate count, or concrete list is visible.
- When a count is uncertain, use careful visible phrasing such as "at least N" or "approximately N" instead of guessing.
- Use viewer perspective consistently for left, right, top, bottom, foreground, background, front, and behind.
- Match the caption strategy to the image type. A photo-like scene and a structured poster, UI screen, slide, or infographic need different caption organization.

Forbidden caption content:
- Do not mention that there is no watermark, no blur, no crop, no occlusion, no corruption, no low quality, no missing content, no recognizable text, no readable text, or no intrusive watermark.
- Do not mention that a subject is "not cropped", "not occluded", "not blurry", "fully visible", "clear", or "unobstructed" unless a visible boundary, occlusion, blur, or crop actually matters to describing what is present.
- Do not explain why text cannot be read, such as because of low resolution, blur, darkness, occlusion, or distance.
- Do not include quality-filter decisions, safety-filter decisions, watermark decisions, or training-value judgments.

Caption content requirements:
- Describe the visible visual medium, image type, and style when apparent, using only visually supported words such as photo, film photo, line drawing, simple sketch, flat vector illustration, watercolor, oil-painting-style artwork, anime-style illustration, 3D render, pixel art, collage, poster design, UI screenshot, slide, product render, e-commerce catalog image, or magazine cover.
- Count the main people, animals, and visually important repeated objects inside the captions when possible.
- For photo-like or natural scenes such as portraits, landscapes, animals, interiors, street scenes, products in real settings, and other real-world scenes, describe the camera or viewing angle when visible or useful, framing, shot distance, perspective and depth, foreground/midground/background layers, background context, depth of field or bokeh if present, lighting, visual style, and the spatial structure of the scene.
- For each visible main person, describe their viewer-relative position, posture, pose, action, gesture, gaze direction, facial expression, hairstyle, skin tone when visible, visible appearance, clothing, accessories, and how their body relates spatially to other people or objects.
- For each main animal or object, describe visible position, shape, color, texture, pose/state, parts, surface markings, attachments, and relation to nearby objects when visible.
- For crowded scenes, distinguish foreground, main subject area, and background crowd. Count the main visible subjects first, then describe the crowd as approximate or uncountable if needed.
- For object, product, UI, poster, slide, design, and infographic images, do not collapse the image into a one-sentence summary. Describe the overall structure and each major visible component concretely.

Layout, overlap, and design requirements:
- For posters, UI screens, slides, e-commerce images, typography-heavy designs, infographics, and composited images, describe the layout professionally and objectively.
- For structured UI, poster, slide, infographic, design, or composited images, first describe the base/background layer as an image: visible people, scenery, objects, colors, lighting, and depth. Then describe the top design layer.
- Describe top design layers in a stable reading order: top-to-bottom and left-to-right, region by region or panel by panel.
- For each visible region, panel, card, header, sidebar, footer, chart, menu, button, badge, icon group, or image block, describe its position, size relationship, background color/shape, icons, images, text blocks, labels, numbers, dividers, and important separators when visible.
- Describe major regions, alignment, visual hierarchy, foreground/background layers, image blocks, text blocks, typography placement, and spatial relationships.
- For text in different regions, state the exact OCR string, location, typography style, approximate size relationship, color, weight, orientation, and how it is aligned or grouped when visible.
- If one element overlaps or covers another, state what is in front and what is behind. If text is partly covered, state which visible text or letters are partly obscured and what foreground element covers them.
- If a person, product, frame, panel, sticker, button, or other layer sits above a text layer, describe that z-order relationship when it affects the visible design.
- Mention crop, partial visibility, or occlusion only when it actually appears in the image and affects the visible composition.

Text requirements:
- All readable OCR text from context.text.visible_text must appear verbatim in short_en, short_zh, long_en, and long_zh.
- Put exact OCR strings in double quotes in every caption.
- Do not invent text.
- If unreadable or garbled text-like areas are visually important, describe only the visible fact, such as "small unreadable newspaper text areas in the background" or "tiny unreadable UI labels near the bottom". Do not explain the reason they are unreadable.
- Unreadable text regions from context are metadata for awareness only. They do not need to appear in captions unless they are visually important to the scene or layout.

Caption requirements:
- Produce short_en, short_zh, long_en, and long_zh.
- Chinese and English captions must be semantically equivalent.
- short captions should be concise but must still preserve all OCR strings.
- long captions should be detailed and objective. Do not be terse when the image contains many people, many objects, much text, or a complex graphic layout.
- For complex natural scenes, long captions should cover camera or viewing angle, perspective/depth, foreground/midground/background structure, visible style, subject details, and spatial relationships when these are visible.
- For structured design images, long captions should describe the base layer first, then the overlay/layout from top-to-bottom and left-to-right, including all readable text, visible controls, icons, blocks, charts, typography, and layering relationships.
- Avoid marketing language, storytelling, aesthetic praise, unsupported interpretation, and data-quality report wording.

Output JSON schema:
{{SCHEMA_JSON}}
\end{promptbox}



\section{RL Post-training Details}
\label{app:rl-post-training-details}

\subsection{DiffusionNFT Objective}
\label{app:diffusionnft-objective}

DiffusionNFT~\citep{zheng2025diffusionnft} optimizes a diffusion generator from final generated samples and rewards, without storing the full reverse denoising trajectory. For each prompt group $G(p)$, rewards are converted into normalized advantages:

\begin{equation} A_i = \frac{r_i - \operatorname{mean}_{j \in G(p)}(r_j)}{\sigma}, \end{equation}

where $r_i$ is the reward of sample $i$, and $\sigma$ is a prompt-level or global reward standard deviation. The advantage is clipped and mapped to a reward-dependent mixing coefficient:

\begin{equation} \rho_i = \operatorname{clip} \left( \frac{\operatorname{clip}(A_i, -A_{\max}, A_{\max})}{2A_{\max}} + \frac{1}{2}, 0, 1 \right). \end{equation}

Given the current prediction $v_{\theta}$ and the frozen old-policy prediction $v_{\mathrm{old}}$ on the same noised latent input, DiffusionNFT constructs a positive prediction and an implicit negative prediction:

\begin{align} v^{+} &= \beta v_{\theta} + (1-\beta)v_{\mathrm{old}}, \ v^{-} &= (1+\beta)v_{\mathrm{old}} - \beta v_{\theta}. \end{align}

The final loss interpolates between positive and implicit-negative losses:

\begin{equation} \mathcal{L}_{\mathrm{NFT}} = \rho_i \mathcal{L}^{+} + (1-\rho_i)\mathcal{L}^{-}. \end{equation}
Therefore, high-reward samples receive larger positive weights, while low-reward samples act as implicit negatives. The old policy anchors the update; in our online setting, it is instantiated by the checkpoint that generated the current batch.

\subsection{RL Training Ablation}
\label{app:rl-training-ablation}

To isolate the effect of RL post-training, we compare the 5B model before and
after this stage. The SeFi-Image-5B entry in Sec.~\ref{sec:performance-evaluation}
uses the w/ RL setting, while
Tables~\ref{tab:rl-ablation-geneval}--\ref{tab:rl-ablation-oneig-en} report both
settings side by side. RL mainly improves text rendering and prompt-following
metrics, with clear gains on LongTextBench, CVTG-2K word accuracy, and OneIG.
Compositional scores remain largely stable, and DPG-Bench shows only a small
overall change.

\begin{table}[H]
  \centering
  \sefitablefont

  \begin{minipage}[t]{0.48\textwidth}
    \centering
    \caption{RL ablation on GenEval.}
    \label{tab:rl-ablation-geneval}
    \resizebox{\linewidth}{!}{
      \begin{tabular}{lccccccc}
        \toprule
        Model & Single Obj. & Two Obj. & Counting & Colors & Position & Attr. Binding & Overall$\uparrow$ \\
        \midrule
        w/ RL & 1.00 & 0.92 & 0.87 & 0.90 & 0.84 & 0.76 & 0.88 \\
        w/o RL & 1.00 & 0.92 & 0.85 & 0.90 & 0.81 & 0.77 & 0.87 \\
        \bottomrule
      \end{tabular}
    }
  \end{minipage}
  \hfill
  \begin{minipage}[t]{0.48\textwidth}
    \centering
    \caption{RL ablation on DPG-Bench.}
    \label{tab:rl-ablation-dpg}
    \resizebox{\linewidth}{!}{
      \begin{tabular}{lcccccc}
        \toprule
        Model & Global & Entity & Attribute & Relation & Other & Overall$\uparrow$ \\
        \midrule
        w/o RL & 93.06 & 92.46 & 91.75 & 92.56 & 90.73 & 87.45 \\
        w/ RL & 88.24 & 92.62 & 91.49 & 93.57 & 91.55 & 87.27 \\
        \bottomrule
      \end{tabular}
    }
  \end{minipage}
\end{table}

\begin{table}[H]
  \centering
  \sefitablefont

  \begin{minipage}[t]{0.48\textwidth}
    \centering
    \caption{RL ablation on LongTextBench.}
    \label{tab:rl-ablation-longtextbench}
    \begin{tabular}{lccc}
      \toprule
      Model & EN$\uparrow$ & ZH$\uparrow$ & Avg$\uparrow$ \\
      \midrule
      w/ RL & 0.9778 & 0.9782 & 0.9780 \\
      w/o RL & 0.9699 & 0.9631 & 0.9665 \\
      \bottomrule
    \end{tabular}
  \end{minipage}
  \hfill
  \begin{minipage}[t]{0.48\textwidth}
    \centering
    \caption{RL ablation on CVTG-2K.}
    \label{tab:rl-ablation-cvtg}
    \begin{tabular}{lccc}
      \toprule
      Model & NED$\uparrow$ & CLIPScore$\uparrow$ & Word Acc.$\uparrow$ \\
      \midrule
      w/ RL & 0.9434 & 0.8155 & 0.8947 \\
      w/o RL & 0.9365 & 0.8202 & 0.8783 \\
      \bottomrule
    \end{tabular}
  \end{minipage}
\end{table}

\begin{table}[H]
  \centering
  \sefitablefont

  \begin{minipage}[t]{0.48\textwidth}
    \centering
    \caption{RL ablation on OneIG-ZH.}
    \label{tab:rl-ablation-oneig-zh}
    \resizebox{\linewidth}{!}{
      \begin{tabular}{lcccccc}
        \toprule
        Model & Alignment & Text & Reasoning & Style & Diversity & Overall$\uparrow$ \\
        \midrule
        w/ RL & 0.8063 & 0.9716 & 0.2672 & 0.4307 & 0.2137 & 0.5379 \\
        w/o RL & 0.8113 & 0.9619 & 0.2648 & 0.4288 & 0.2008 & 0.5335 \\
        \bottomrule
      \end{tabular}
    }
  \end{minipage}
  \hfill
  \begin{minipage}[t]{0.48\textwidth}
    \centering
    \caption{RL ablation on OneIG-EN.}
    \label{tab:rl-ablation-oneig-en}
    \resizebox{\linewidth}{!}{
      \begin{tabular}{lcccccc}
        \toprule
        Model & Alignment & Text & Reasoning & Style & Diversity & Overall$\uparrow$ \\
        \midrule
        w/ RL & 0.8648 & 0.9576 & 0.3240 & 0.4502 & 0.2065 & 0.5606 \\
        w/o RL & 0.8703 & 0.9428 & 0.3111 & 0.4491 & 0.1973 & 0.5541 \\
        \bottomrule
      \end{tabular}
    }
  \end{minipage}
\end{table}

\section{Turbo Model Performance}
\label{app:distillation-details}

We report the performance of the 4-step turbo variants distilled from the SFT checkpoint via DMD2 (Sec.~\ref{subsec:training-distillation}). As shown in the benchmark tables below, step compression introduces a modest and consistent quality trade-off: SeFi-Image-5B-Turbo typically loses roughly 1--4 points relative to its full-step teacher across benchmarks, yet matches or outperforms Z-Image-Turbo on most evaluation axes despite substantially lower training compute. The degradation is smallest on compositional tasks (GenEval: 0.86 vs. 0.87; DPG: 86.45 vs. 87.45), likely because the semantic branch commits high-level structure early in the reverse process, leaving less residual work for the removed intermediate steps. The gap widens on text-heavy benchmarks such as LongTextBench and CVTG-2K, where fine-grained character rendering benefits from additional denoising iterations. Notably, certain sub-metrics, particularly Style on OneIG and CLIPScore on CVTG-2K, are preserved or even slightly improved in the turbo models, suggesting that the distilled trajectory retains sufficient capacity for global aesthetics and text-image alignment.

\begin{figure}[t]
  \centering
  \includegraphics[width=0.7\linewidth]{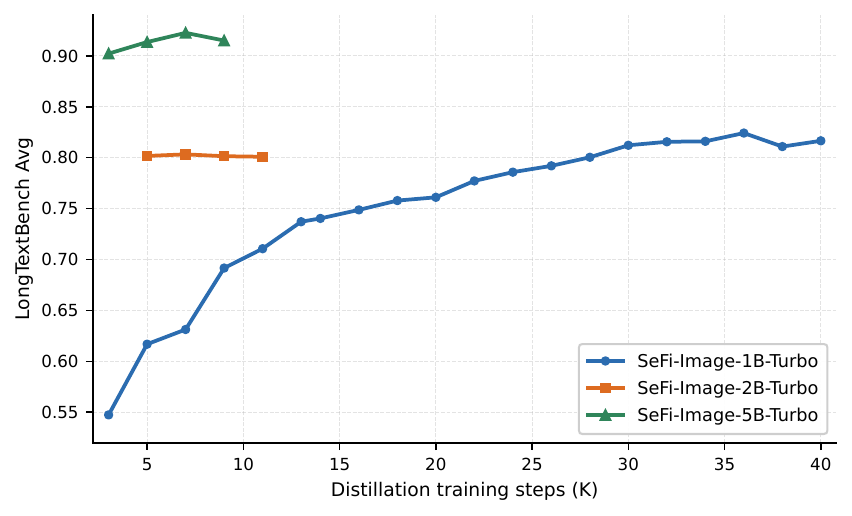}
  \caption{LongTextBench Avg during DMD2 distillation for different model scales. The 1B curve combines an initial 3K--13K run and a continued run; the continued run's logged 1K checkpoint is plotted as 14K total steps.}
  \label{fig:turbo-longtext-convergence}
\end{figure}

During distillation training, we observe that smaller models require significantly more iterations to converge. Figure~\ref{fig:turbo-longtext-convergence} shows LongTextBench Avg scores along training: the 1B variant keeps improving well beyond 13K steps and becomes stable only around the 30K-step range, whereas the 2B and 5B variants converge after roughly 5K steps. We attribute this to larger models starting closer to the target teacher distribution, enabling the fake-score and teacher-score networks to estimate the distribution gap more accurately from the outset. We also note that SeFi-Image-1B-Turbo outperforms its 2B-Turbo counterpart on several benchmarks, such as GenEval, LongTextBench, and OneIG. This inversion mirrors the pattern already observed between their full-step teachers in Sec.~\ref{sec:performance-evaluation} and likely reflects capacity-dependent language and capability trade-offs.

\begin{table}[H]
  \centering
  \sefitablefont

  \begin{minipage}[t]{0.48\textwidth}
    \centering
    \caption{LongTextBench results.}
    \label{tab:turbo-longtextbench-performance}
    \resizebox{\linewidth}{!}{%
      \begin{tabular}{lccc}
        \toprule
        Model & EN$\uparrow$ & ZH$\uparrow$ & Avg$\uparrow$ \\
        \midrule
        SeFi-Image-5B & 0.970 & 0.963 & 0.967 \\
        SeFi-Image-5B-Turbo & 0.903 & 0.942 & 0.922 \\
        Z-Image-Turbo & 0.917 & 0.926 & 0.921 \\
        SeFi-Image-1B & 0.854 & 0.856 & 0.855 \\
        SeFi-Image-2B & 0.870 & 0.824 & 0.847 \\
        SeFi-Image-1B-Turbo & 0.825 & 0.824 & 0.824 \\
        SeFi-Image-2B-Turbo & 0.799 & 0.803 & 0.801 \\
        \bottomrule
      \end{tabular}%
    }
  \end{minipage}
  \hfill
  \begin{minipage}[t]{0.48\textwidth}
    \centering
    \caption{CVTG-2K results.}
    \label{tab:turbo-cvtg-performance}
    \resizebox{\linewidth}{!}{%
      \begin{tabular}{lccc}
        \toprule
        Model & NED$\uparrow$ & CLIPScore$\uparrow$ & Word Acc.$\uparrow$ \\
        \midrule
        SeFi-Image-5B & 0.937 & 0.820 & 0.878 \\
        Z-Image-Turbo & 0.928 & 0.805 & 0.859 \\
        SeFi-Image-5B-Turbo & 0.899 & 0.833 & 0.782 \\
        SeFi-Image-2B & 0.896 & 0.816 & 0.773 \\
        SeFi-Image-2B-Turbo & 0.878 & 0.821 & 0.712 \\
        SeFi-Image-1B & 0.862 & 0.810 & 0.718 \\
        SeFi-Image-1B-Turbo & 0.859 & 0.818 & 0.707 \\
        \bottomrule
      \end{tabular}%
    }
  \end{minipage}
\end{table}

\begin{table}[H]
  \centering
  \caption{GenEval results for turbo variants.}
  \label{tab:turbo-geneval-performance}
  \sefitablefont
  \begin{tabular}{lccccccc}
    \toprule
    Model & Single Obj. & Two Obj. & Counting & Colors & Position & Attr. Binding & Overall$\uparrow$ \\
    \midrule
    SeFi-Image-5B & 1.00 & 0.92 & 0.85 & 0.90 & 0.81 & 0.77 & 0.87 \\
    SeFi-Image-2B & 0.99 & 0.93 & 0.84 & 0.91 & 0.78 & 0.78 & 0.87 \\
    SeFi-Image-1B & 0.99 & 0.91 & 0.83 & 0.92 & 0.82 & 0.75 & 0.87 \\
    SeFi-Image-5B-Turbo & 0.99 & 0.93 & 0.81 & 0.90 & 0.84 & 0.71 & 0.86 \\
    SeFi-Image-1B-Turbo & 0.98 & 0.90 & 0.70 & 0.86 & 0.86 & 0.74 & 0.84 \\
    SeFi-Image-2B-Turbo & 0.98 & 0.91 & 0.66 & 0.91 & 0.79 & 0.73 & 0.83 \\
    Z-Image-Turbo & 1.00 & 0.95 & 0.77 & 0.89 & 0.65 & 0.68 & 0.82 \\
    \bottomrule
  \end{tabular}
\end{table}

\begin{table}[H]
  \centering
  \caption{DPG-Bench results for turbo variants.}
  \label{tab:turbo-dpg-performance}
  \sefitablefont
  \begin{tabular}{lcccccc}
    \toprule
    Model & Global & Entity & Attribute & Relation & Other & Overall$\uparrow$ \\
    \midrule
    SeFi-Image-5B & 93.06 & 92.46 & 91.75 & 92.56 & 90.73 & 87.45 \\
    SeFi-Image-2B & 89.44 & 91.81 & 92.02 & 93.04 & 92.13 & 87.31 \\
    SeFi-Image-1B & 91.19 & 93.16 & 91.59 & 92.13 & 86.66 & 87.17 \\
    SeFi-Image-5B-Turbo & 85.85 & 91.27 & 92.42 & 90.88 & 91.61 & 86.45 \\
    SeFi-Image-2B-Turbo & 90.36 & 92.01 & 90.15 & 93.60 & 92.14 & 86.14 \\
    SeFi-Image-1B-Turbo & 91.24 & 91.21 & 91.56 & 91.39 & 89.71 & 85.34 \\
    Z-Image-Turbo & 91.29 & 89.59 & 90.14 & 92.16 & 88.68 & 84.86 \\
    \bottomrule
  \end{tabular}
\end{table}

\begin{table}[H]
  \centering
  \caption{OneIG results for turbo variants.}
  \label{tab:turbo-oneig-performance}
  \sefitablefont

  \begin{subtable}[t]{\textwidth}
    \centering
    \caption{OneIG-ZH}
    \label{tab:turbo-oneig-zh-performance}
    \begin{tabular}{lcccccc}
      \toprule
      Model & Alignment & Text & Reasoning & Style & Diversity & Overall$\uparrow$ \\
      \midrule
      SeFi-Image-5B & 0.8113 & 0.9619 & 0.2648 & 0.4288 & 0.2008 & 0.5335 \\
      SeFi-Image-5B-Turbo & 0.8051 & 0.9287 & 0.2536 & 0.4402 & 0.1596 & 0.5174 \\
      SeFi-Image-1B & 0.7920 & 0.9022 & 0.2394 & 0.3934 & 0.2206 & 0.5095 \\
      Z-Image-Turbo & 0.7820 & 0.9820 & 0.2760 & 0.3610 & 0.1340 & 0.5060 \\
      SeFi-Image-1B-Turbo & 0.7660 & 0.8754 & 0.2324 & 0.4136 & 0.2150 & 0.5005 \\
      SeFi-Image-2B & 0.7974 & 0.8120 & 0.2476 & 0.4043 & 0.2280 & 0.4979 \\
      SeFi-Image-2B-Turbo & 0.7924 & 0.7919 & 0.2376 & 0.4007 & 0.1714 & 0.4788 \\
      \bottomrule
    \end{tabular}
  \end{subtable}

  \vspace{0.8em}

  \begin{subtable}[t]{\textwidth}
    \centering
    \caption{OneIG-EN}
    \label{tab:turbo-oneig-en-performance}
    \begin{tabular}{lcccccc}
      \toprule
      Model & Alignment & Text & Reasoning & Style & Diversity & Overall$\uparrow$ \\
      \midrule
      SeFi-Image-5B & 0.8703 & 0.9428 & 0.3111 & 0.4491 & 0.1973 & 0.5541 \\
      SeFi-Image-2B & 0.8664 & 0.9017 & 0.2794 & 0.4209 & 0.2150 & 0.5367 \\
      Z-Image-Turbo & 0.8400 & 0.9940 & 0.2980 & 0.3680 & 0.1390 & 0.5280 \\
      SeFi-Image-1B & 0.8594 & 0.8839 & 0.2706 & 0.3975 & 0.2121 & 0.5247 \\
      SeFi-Image-5B-Turbo & 0.8582 & 0.8557 & 0.2904 & 0.4529 & 0.1584 & 0.5231 \\
      SeFi-Image-1B-Turbo & 0.8289 & 0.8242 & 0.2849 & 0.4258 & 0.2025 & 0.5133 \\
      SeFi-Image-2B-Turbo & 0.8582 & 0.8251 & 0.2685 & 0.4165 & 0.1621 & 0.5061 \\
      \bottomrule
    \end{tabular}
  \end{subtable}
\end{table}

\section{Extend to Higher Resolution}
\label{app:higher-resolution}

\begin{figure}[H]
  \centering
  \includegraphics[width=0.75\linewidth]{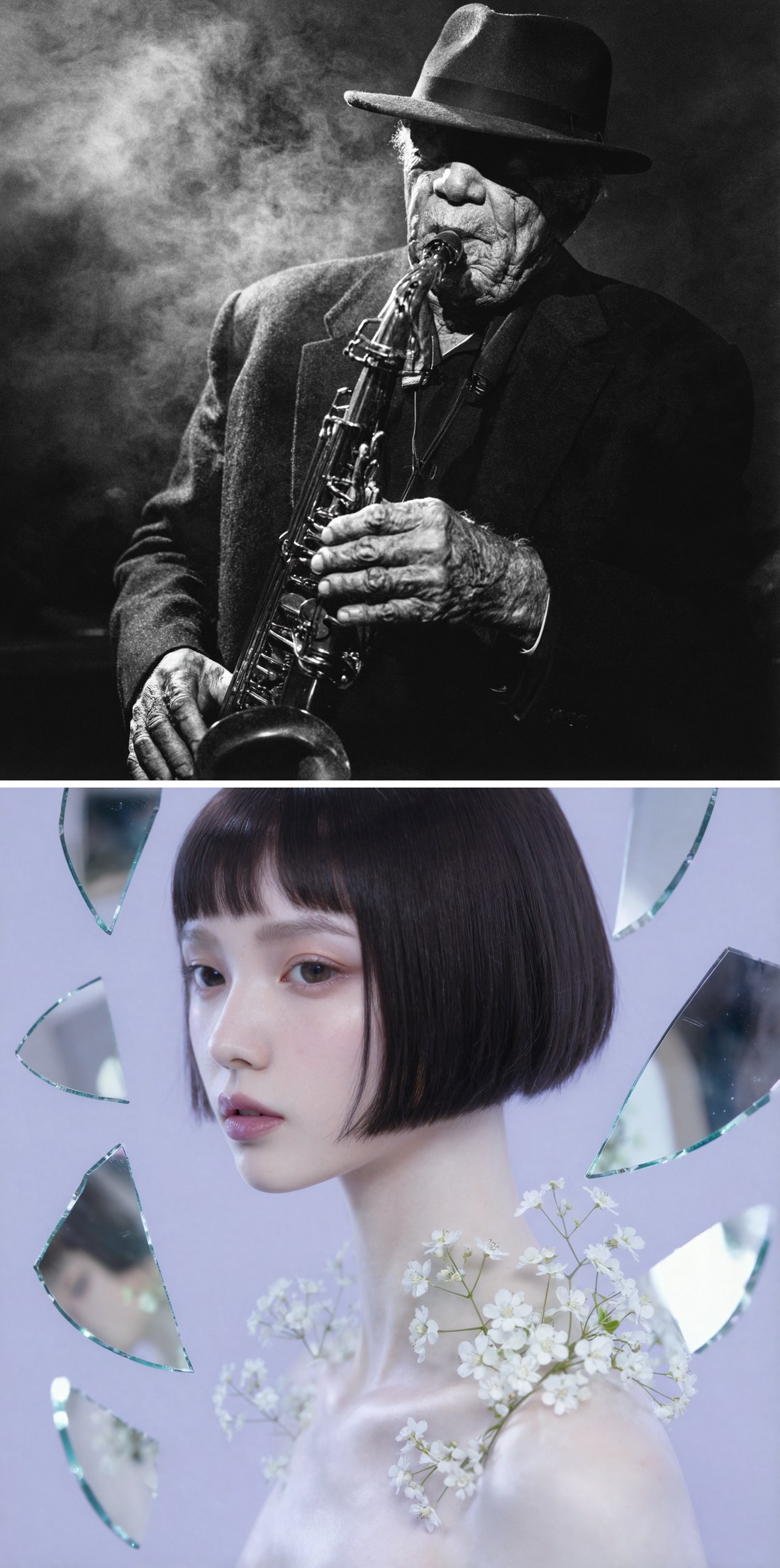}
  \caption{Training-free 1440px samples generated by the 5B version of \method.}
  \label{fig:higher-resolution-1440px-a}
\end{figure}

\begin{figure}[H]
  \centering
  \includegraphics[width=0.75\linewidth]{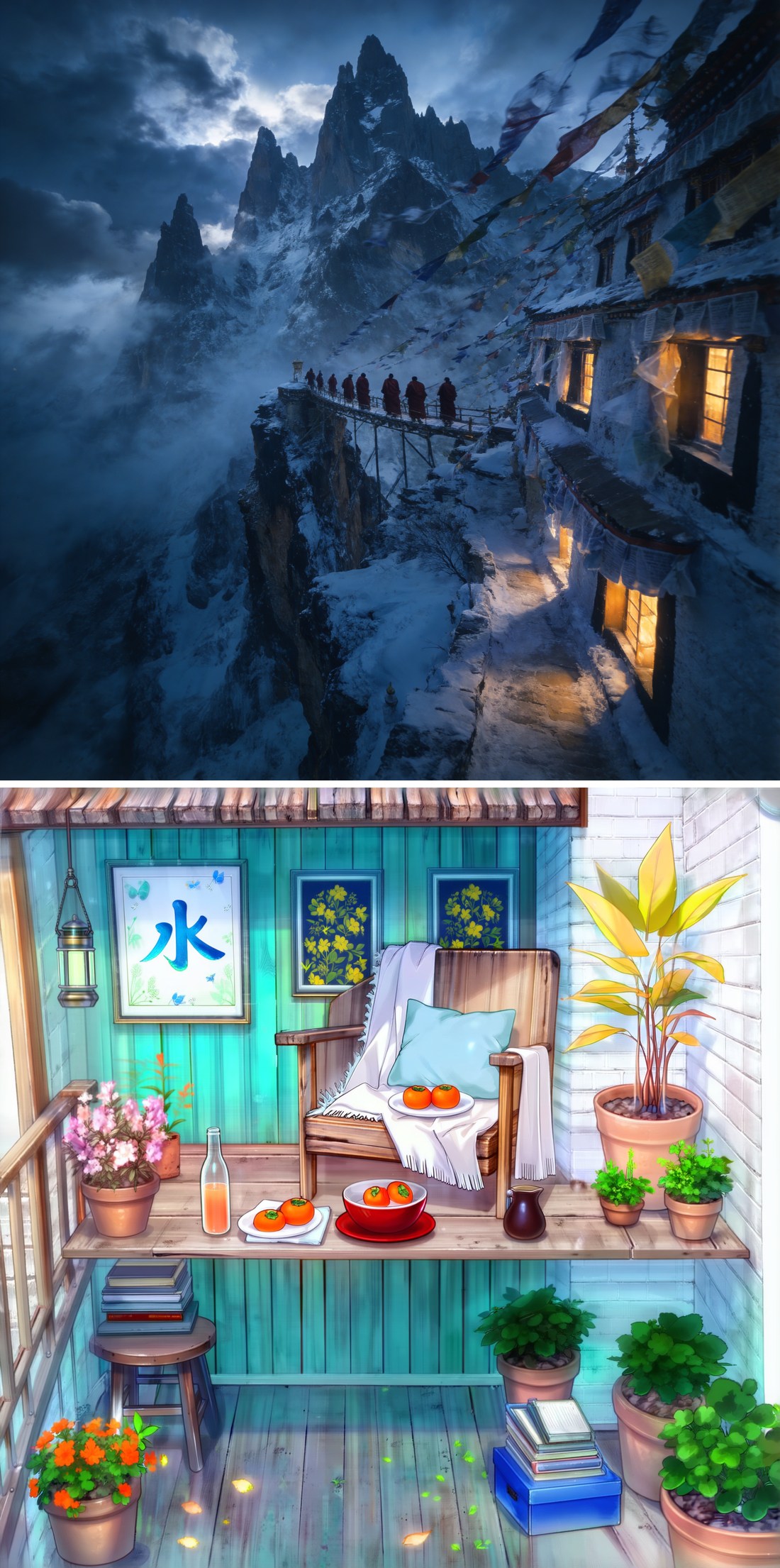}
  \caption{Additional training-free 1440px samples generated by the 5B version of \method.}
  \label{fig:higher-resolution-1440px-b}
\end{figure}

Although \method is trained at 1024px resolution, the same checkpoint can also
generate higher-resolution images at inference time without additional training.
In practice, we directly increase the generation canvas to 1440px while keeping
the model weights and sampling recipe unchanged. Figures~\ref{fig:higher-resolution-1440px-a}
and~\ref{fig:higher-resolution-1440px-b} show qualitative 1440px samples produced
in this training-free setting. The examples suggest that \method can preserve
coherent global composition and local visual detail beyond its training
resolution, while systematic high-resolution benchmarking is left for future work.

\clearpage
